# Strapdown Inertial Navigation System Initial Alignment based on Group of Double Direct Spatial Isometries

Lubin Chang, Fangjun Qin, and Jiangning Xu

***Abstract*—The task of strapdown inertial navigation system (SINS) initial alignment is to calculate the attitude transformation matrix from body frame to navigation frame. In this paper, such attitude transformation matrix is divided into two parts through introducing the initial inertially fixed navigation frame as inertial frame. The attitude changes of the navigation frame corresponding to the defined inertial frame can be exactly calculated with known velocity and position provided by GNSS. The attitude from body frame to the defined inertial frame is estimated based on the SINS mechanization in inertial frame. The attitude, velocity and position in inertial frame are formulated together as element of the group of double direct spatial isometries ( $\mathbb{SE}_2(3)$ ).It is proven that the group state model in inertial frame satisfies a particular "group affine" property and the corresponding error model satisfies a "log-linear" autonomous differential equation on the Lie algebra. Based on such striking property, the attitude from body frame to the defined inertial frame can be estimated based on the linear error model with even extreme large misalignments. Two different error state vectors are extracted based on $\mathbb{SE}_2(3)$ right and left matrix multiplications and the detailed linear state-space models are derived based on the right and left errors for SINS mechanization in inertial frame. With the derived linear state-space models, the explicit initial alignment procedures have been presented. Extensive simulation and field tests indicate that the initial alignment based on the left error model can perform quite well within a wide range of initial attitude errors, although the used filter is still a type of linear Kalman filter. This method is promising in practical products abandoning the traditional coarse alignment stage.**

## I. Introduction

THE strapdown inertial navigation system (SINS) equipped with accelerometers and gyroscopes can provide attitude, velocity and position successively in a self-contained manner [1]. Initial alignment is the essential stage for the dead-reckoning based SINS. The initial velocity and position can be easily obtained from other navigation sources, say GNSS. In contrast, finding the initial attitude is much tricky as no sensor can measure it accurately and directly. In this respect, the initial alignment always refers in particular to attitude alignment. The rapidity and accuracy are the ultimate pursuit for SINS initial alignment, although they are contradicted with each other [2].

Traditionally, there are two successive alignment stages, i.e. coarse alignment and fine alignment. In the coarse alignment stage, some analytical methods are applied making use of the relationship between the inertial sensors measurements and the geographical information [1, 2]. The fine alignment, always based on state estimation methods, such as Kalman filtering (KF), is used to refine the coarse alignment results [3]. Meanwhile, the inertial sensors drift biases are expected to be estimated in the fine alignment. Since the traditional coarse alignment is only suitable under static condition, much effort has been made to seek robust coarse alignments which are suitable for more practical situations, such as swaying mooring condition or even in-flight condition. In recent few years, the most representative coarse alignment method is the optimization-based alignment (OBA). The OBA transforms the initial alignment problem into a constant attitude determination problem using vector observations [4-8]. It is more robust to angular disturbance and noise than traditional coarse alignment. Aiding by GNSS information, it can also perform in-flight alignment. Given the roughly known attitude by the OBA, the KF-based fine alignment can be readily carried out. A key issue affect the fine alignment performance is the state observability. Different state model, maneuverability or measurements can result in different state observability. In recent few years, Silva et al. have investigated the KF-based fine alignment from the perspective of observability/estimability, measurement selection and estimation algorithms [9-11].

There will be inevitable transitions between the coarse and fine alignment stages. Meanwhile, there is no definite time duration for coarse alignment to provide adequate accurate result for the following fine alignment. Different application conditions necessitate different time allocation for the two stages. With this consideration, researchers have never given up to seek unified alignment methods which can abandon the traditional the two-stage based procedures. The first effort for such end is making use of nonlinear filtering based on the nonlinear error state models [12-16]. However, due to the defects of the nonlinear error state models, such methods

The paper was supported in part by National Natural Science Foundation of China (61873275).

The authors are all with the Department of Navigation Engineering, Naval University of Engineering. (e-mail: changlubin@163.com, haig2005@126.com, alimoon@aliyun.com).

cannot reach the same accuracy as the traditional two-stage method. Meanwhile, nonlinear filtering will take much more computational resource than linear KF. Moreover, observability problem encountered by KF will also affect the nonlinear filtering.

In recent few years, the matrix Lie group theory has been applied in some inertial based applications to circumvent the "*false observability*" problem. These applications include simultaneous localization and mapping (SLAM), visual inertial odometry (VIO), attitude estimation and so on [17-26]. Actually, matrix Lie group theory has been applied for long in inertial navigation. This is essentially a matter of representation regarding orientation, and how to treat rotation errors in a filter. In the traditional fine alignment, the Euler angles can be viewed as the rotation errors corresponding to the matrix Lie group $\mathbb{SO}(3)$. It is pointed out that for some inertial based dynamic models, if the state is defined on a Lie group, the dynamic models satisfy a particular "*group affine*" property. Based on such property, the corresponding linear error state model can reflect the nonlinear group error propagation process exactly. It has been found that inertial navigation equations on flat earth are group-affine, that is, they possess some kind of linearity when using a judicious Lie group embedding. The Lie group in question that makes the equations group affine was found to be the group of double direct isometries ( $\mathbb{SE}_2(3)$ ) [25-28]. Based on this knowledge, the invariant extended Kalman filtering (IEKF) is proposed to perform the unified initial alignment without coarse alignment [25, 26]. Commercial navigation systems applications indicate that the IEKF performs quite well in initial alignment even with very large initial misalignments. Moreover, it is still a type of linear KF and the computational cost is similar with the traditional KF based methods. In [26], the initial alignment is used as an industrial application example of the IEKF. The SINS differential equations on flat earth are virtually expressed in inertial reference frame, which is not suitable for navigation on earth surface. Even in the inertial frame system model, there is either no technical detail about the state-space model and filtering procedure. For SINS applications on earth surface, the local-level geographical frame is commonly used as the navigation frame, since the user-intuitive attitude and position can be obtained directly from the output of the mechanization equations. For this mechanization, there is a tightly coupling between the navigation parameters due to the earth rotation and movement of the local-level frame over Earth's curvature. Unfortunately, SINS mechanization in the local-level frame does not satisfy the group affine property due to the strong coupling between the attitude, velocity and position. In previous works, the earth rotation and movement of the local-level frame over Earth's curvature have been ignored in most applications of the IEKF as pointed out in [27]. These facts mainly motivate this paper, which is devoted to judicious making use of the matrix Lie group theory for SINS initial alignment with consideration of the earth rotation and movement of the local-level frame over Earth's curvature. Based on the aforementioned discussion, this paper introduces an inertial frame which is artificial defined through freezing the local-level navigation frame at the very start of the initial alignment task. Based on the defined inertial frame, the attitude from body frame to navigation frame can be divided into two parts. One encodes the attitude changes of the navigation frame corresponding to its initial position. This part can be calculated directly if the geographical position and velocity are known, which is always the case for initial alignment aided by GNSS. Another part of the decomposed attitude represents the attitude from body frame to inertial frame (artificially defined). The kinetics of this attitude is just the attitude differential equation in inertial frame. Based on the calculated attitude from navigation frame to the defined inertial frame and the known velocity and position, we can provide the initial values and inputs for the velocity and position differential equations in the defined inertial frame. If the attitude, velocity and position in the defined inertial frame are defined on $\mathbb{SE}_2(3)$ , the corresponding dynamic model satisfies the group affine property. Therefore, the attitude from body frame to the defined inertial frame can be estimated using the corresponding linear error model with even very large attitude misalignments. After obtaining the two attitude parts, the desired attitude from body frame to navigation frame can be readily obtained, that is, the initial alignment is accomplished.

It should be noted that the $\mathbb{SE}_2(3)$ tool is mainly used to define the group state errors and derive the corresponding vector state error models in this paper. Making use of Lie group to design the SINS calculation algorithm is not the research topic of this paper, as that in [29]. That is to say, in our investigated methods, SINS calculation is still accomplished using the traditional methods. In [30, 31], the special orthogonal group $\mathbb{SO}(3)$ is used to represent the initial attitude matrix in the framework of OBA method, which can improve the convergent speed. Just as we have pointed out that, the OBA is essentially a coarse alignment method. In contrast, the presented method is a linear KF based alignment method using ingeniously defined state errors, which is expected to unify the coarse and fine alignment even with extreme large misalignments. That is to say the alignment structure of the presented method is totally different with that in [30, 31].

The remainder of the paper is organized as follows. Section II presents the mathematical tools that we will use to derive the error state model for $\mathbb{SE}_2(3)$ -KF. In Section III, the desired attitude by initial alignment is decomposed into two parts through introducing the initial inertially fixed navigation frame. In section IV, the error state models are presented for SINS mechanization in inertia frame through formulation the attitude, velocity and position as $\mathbb{SO}(3)+\mathbb{R}^6$ . Section V derives the detailed error state models based on $\mathbb{SE}_2(3)$ theory. Both right and left error definitions are considered in this section. Section VI shows the explicit application procedures of the $\mathbb{SE}_2(3)$ -KF for initial alignment. In section VII, the sound theoretical basis for the reason why the linear error state model can be used to accomplish the initial alignment with large misalignments is presented. Section VIII carries out extensive simulations to evaluate the performance of the $\mathbb{SE}_2(3)$ -KF based initial alignment. Section IX is further devoted to evaluating the performance of these linear initial alignment methods making use of field test data. Conclusions are drawn in the final section.

## II. Mathematical Preliminaries

In this section, we will present the bare minimum of matrix Lie group and the associated Lie algebra which are required to derive the error state model for the $\mathbb{SE}_2(3)$ -KF. For more details about the Lie group theory, one can refer to [26, 32].

The group of double direct spatial isometries introduced in [26, 28] is given by

$$\mathbb{SE}_2(3) = \left\{ \mathbf{T} = \left[ \begin{array}{c|cc} \mathbf{C} & \mathbf{v} & \mathbf{p} \\ \hline \mathbf{0}_{2\times3} & \multicolumn{2}{c}{\mathbf{I}_{2\times2}} \end{array} \right] \in \mathbb{R}^{5\times5} \middle| \begin{array}{l} \mathbf{C} \in \mathbb{SO}(3) \\ \mathbf{v}, \mathbf{p} \in \mathbb{R}^3 \end{array} \right\} \quad (1)$$

where $\mathbb{SO}(3)$ is the special orthogonal group. The inverse of an element of $\mathbb{SE}_2(3)$ is also element of $\mathbb{SE}_2(3)$ . Given the matrix Lie group in (1), the corresponding inverse is given by

$$\mathbf{T}^{-1} = \left[ \begin{array}{c|cc} \mathbf{C}^T & -\mathbf{C}^T\mathbf{v} & -\mathbf{C}^T\mathbf{p} \\ \hline \mathbf{0}_{2\times3} & \multicolumn{2}{c}{\mathbf{I}_{2\times2}} \end{array} \right] \in \mathbb{SE}_2(3) \quad (2)$$

The Lie algebra associated with $\mathbb{SE}_2(3)$ is given by

$$\mathfrak{se}_2(3) = \left\{ \boldsymbol{\Xi} = \boldsymbol{\zeta}^\wedge \in \mathbb{R}^{5\times5} \middle| \boldsymbol{\zeta} \in \mathbb{R}^9 \right\} \quad (3)$$

where

$$\boldsymbol{\zeta}^\wedge = \begin{bmatrix} \boldsymbol{\varphi} \\ \boldsymbol{\upsilon} \\ \boldsymbol{\rho} \end{bmatrix}^\wedge = \left[ \begin{array}{c|cc} \boldsymbol{\varphi}\times & \boldsymbol{\upsilon} & \boldsymbol{\rho} \\ \hline \mathbf{0}_{2\times3} & \multicolumn{2}{c}{\mathbf{0}_{2\times2}} \end{array} \right] \in \mathbb{R}^{5\times5}, \quad \boldsymbol{\upsilon}, \boldsymbol{\rho} \in \mathbb{R}^3 \quad (4)$$

The matrix Lie group $\mathbb{SE}_2(3)$ is related to its associated Lie algebra $\mathfrak{se}_2(3)$ through the exponential map

$$\mathbf{T} = \exp(\boldsymbol{\zeta}) = \exp_m\left(\boldsymbol{\zeta}^\wedge\right) = \sum_{k=0}^{\infty} \frac{1}{k!}\left(\boldsymbol{\zeta}^\wedge\right)^k \quad (5)$$

The closed-form solution of (5) is given by

$$\mathbf{T} = \exp(\boldsymbol{\zeta}) = \left[ \begin{array}{c|cc} \exp(\boldsymbol{\varphi}) & \mathbf{J}_l\boldsymbol{\upsilon} & \mathbf{J}_l\boldsymbol{\rho} \\ \hline \mathbf{0}_{2\times3} & \multicolumn{2}{c}{\mathbf{I}_{2\times2}} \end{array} \right] \quad (6)$$

where

$$\exp(\boldsymbol{\varphi}) = \cos\varphi \mathbf{I}_{3\times3} + (1-\cos\varphi)\mathbf{a}\mathbf{a}^T + \sin\varphi(\mathbf{a}\times) \quad (7)$$

$$\mathbf{J}_l = \frac{\sin\varphi}{\varphi}\mathbf{I}_{3\times3} + \left(1 - \frac{\sin\varphi}{\varphi}\right)\mathbf{a}\mathbf{a}^T + \left(\frac{1-\cos\varphi}{\varphi}\right)(\mathbf{a}\times) \quad (8)$$

$\boldsymbol{\varphi} = \varphi\mathbf{a}$ where $\varphi$ is the angle of the rotation and $\mathbf{a}$ is the unit-length axis of rotation.

The inverse operation of the exponential map is denoted by

$$\boldsymbol{\zeta} = \ln(\mathbf{T})^\vee \quad (9)$$

Give the matrix Lie group in (1), the solution of (9) is given by

$$\boldsymbol{\zeta} = \begin{bmatrix} \varphi\mathbf{a} \\ \mathbf{J}_l^{-1}\mathbf{v} \\ \mathbf{J}_l^{-1}\mathbf{p} \end{bmatrix} \quad (10)$$

where

$$\varphi = \cos^{-1}\left(\frac{\mathrm{tr}(\mathbf{C})-1}{2}\right), \mathbf{C}\mathbf{a} = \mathbf{a} \quad (11)$$

$$\mathbf{J}_l^{-1} = \frac{\varphi}{2}\frac{\cot\varphi}{2}\mathbf{I}_{3\times3} + \left(1 - \frac{\varphi}{2}\frac{\cot\varphi}{2}\right)\mathbf{a}\mathbf{a}^T - \frac{\varphi}{2}(\mathbf{a}\times) \quad (12)$$

Eq. (11) implies that $\mathbf{a}$ is a (unit-length) eigenvector of $\mathbf{C}$ corresponding to an eigenvalue of 1.

## III. SINS Model Decomposition

In this section, the traditional SINS formulation on navigation frame is decomposed to derive the formulation on inertial frame. The application of SINS formulation on inertial frame aims to reduce the complexity of the corresponding error state equations. Moreover, it will be demonstrated in Section VI that the SINS formulation on inertial frame satisfies a particular group affine properties, which is the theoretical basis for the linear initial alignment with arbitrary initial misalignments.

Denote the body frame by $b$ , the local level navigation frame as $n$ , the earth frame as $e$ and the inertial frame as $i$ . The directions of the used navigation frame are “East-North-Up”. Denote the attitude as the transformation matrix from body frame to navigation frame, i.e. $\mathbf{C}_b^n$ , the velocity vector as $\mathbf{v}^n = \begin{bmatrix} \mathbf{v}_E^n & \mathbf{v}_N^n & \mathbf{v}_U^n \end{bmatrix}^T$ and the position vector as $\mathbf{p}^n = \begin{bmatrix} L & \lambda & h \end{bmatrix}^T$ with $L$ being the latitude, $\lambda$ longitude and $h$ height. The navigation frame differential equations of attitude, velocity and position are given by

$$\begin{bmatrix} \dot{\mathbf{C}}_b^n \\ \dot{\mathbf{v}}^n \\ \dot{\mathbf{p}}^n \end{bmatrix} = \begin{bmatrix} \mathbf{C}_b^n\left(\boldsymbol{\omega}_{nb}^b\times\right) \\ \mathbf{C}_b^n\mathbf{f}^b - \left(2\boldsymbol{\omega}_{ie}^n + \boldsymbol{\omega}_{en}^n\right)\times\mathbf{v}^n + \mathbf{g}^n \\ \mathbf{R}_c\mathbf{v}^n \end{bmatrix} \quad (13)$$

where

$$\boldsymbol{\omega}_{nb}^b = \boldsymbol{\omega}_{ib}^b - \mathbf{C}_b^{nT}\boldsymbol{\omega}_{in}^n = \boldsymbol{\omega}_{ib}^b - \mathbf{C}_b^{nT}\left(\boldsymbol{\omega}_{ie}^n + \boldsymbol{\omega}_{en}^n\right) \quad (14)$$

$\boldsymbol{\omega}_{ib}^b$ is angular rate of the body frame with respect to the navigation frame, expressed in body frame. $\boldsymbol{\omega}_{ib}^b$ can be measured by gyroscopes. $\boldsymbol{\omega}_{ie}^n$ is the earth angular rate expressed in navigation frame and is given by

$$\boldsymbol{\omega}_{ie}^n = \begin{bmatrix} 0 & \omega_{ie}\cos L & \omega_{ie}\sin L \end{bmatrix}^T \quad (15)$$

$\omega_{ie}$ is the earth rotation rate. $\boldsymbol{\omega}_{en}^n$ is the angular rate of the navigation with respect to earth frame, expressed in navigation frame. $\boldsymbol{\omega}_{en}^n$ is caused by the linear motion on the curve of spheroid and is given by

$$\boldsymbol{\omega}_{en}^n = \begin{bmatrix} -\dfrac{\mathbf{v}_N^n}{R_M + h} & \dfrac{\mathbf{v}_E^n}{R_N + h} & \dfrac{\mathbf{v}_E^n\tan L}{R_N + h} \end{bmatrix} \quad (16)$$

where $R_M$ is the meridian radius of curvature of the WGS-84 reference ellipsoid and $R_N$ is the transverse radius. $\mathbf{f}^b$ is the specific force and is measured by accelerometers. $\mathbf{g}^n$ is the gravity vector. $\mathbf{R}_c$ is the local curvature matrix and is given by

$$\mathbf{R}_c = \begin{bmatrix} 0 & \frac{1}{R_M + h} & 0 \\ \frac{\sec L}{R_N + h} & 0 & 0 \\ 0 & 0 & 1 \end{bmatrix} \tag{17}$$

According to the attitude matrix multiplication chain rule, the attitude matrix at time instant $t$ can be decomposed as

$$\mathbf{C}_b^n(t) = \mathbf{C}_{b(t)}^{n(t)} = \mathbf{C}_{n(0)}^{n(t)}\mathbf{C}_{b(t)}^{n(0)} = \mathbf{C}_i^{n(t)}\mathbf{C}_{b(t)}^i \tag{18}$$

where $n(0)$ is the navigation frame at the very state of the initial alignment. $n(0)$ can be viewed as inertial frame because that it is a non-rotational frame. The differential equation of $\mathbf{C}_{n(t)}^i$ is given by

$$\dot{\mathbf{C}}_{n(t)}^i = \mathbf{C}_{n(t)}^i\left(\boldsymbol{\omega}_{in}^n \times\right) \tag{19}$$

According to the definition of the frame $i$, it can be easily deduced that the initial value of $\mathbf{C}_{n(t)}^i$ is error-free, as $\mathbf{C}_{n(0)}^i = \mathbf{C}_{n(0)}^{n(0)} = \mathbf{I}_{3\times3}$. The initial alignment is always carried out aided by the GNSS which can provide precise velocity and position information. Therefore, $\boldsymbol{\omega}_{in}^n$ can also be viewed as known value. In this respect, $\mathbf{C}_{n(t)}^i$ can be readily calculated based on the known velocity and position information with exact initial value, and therefore can be viewed as known value.

The differential equation of $\mathbf{C}_{b(t)}^i$ is given by

$$\dot{\mathbf{C}}_{b(t)}^i = \mathbf{C}_{b(t)}^i\left(\boldsymbol{\omega}_{ib}^b \times\right) \tag{20}$$

It can be deduced that the initial value of $\mathbf{C}_{b(t)}^i$ has a constant misalignment matrix according to the following attitude matrix decomposition

$$\mathbf{C}_{b(t)}^i = \mathbf{C}_{b(t)}^{n(0)} = \mathbf{C}_{b(0)}^{n(0)}\mathbf{C}_{b(t)}^{b(0)} \tag{21}$$

where $b(0)$ is the body frame at the very start of the initial alignment and can also be viewed as inertial frame. The attitude matrix transformation $\mathbf{C}_{b(0)}^{n(0)}$ between two inertial frames is therefore constant.

According to the attitude matrix decomposition, if $\mathbf{C}_{b(t)}^i$ can be estimated, $\mathbf{C}_b^n$ can be readily obtained with the calculated $\mathbf{C}_{n(t)}^i$. Therefore, the problem of estimating $\mathbf{C}_b^n$ has been transformed into estimating $\mathbf{C}_{b(t)}^i$. Accordingly, the inertial frame differential equations of attitude, velocity and position are given by

$$\begin{bmatrix} \dot{\mathbf{C}}_b^i \\ \dot{\mathbf{v}}^i \\ \dot{\mathbf{r}}^i \end{bmatrix} = \begin{bmatrix} \mathbf{C}_b^i\boldsymbol{\omega}_{ib}^b \times \\ \mathbf{C}_b^i\mathbf{f}^b + \mathbf{C}_n^i\mathbf{g}^n \\ \mathbf{v}^i \end{bmatrix} \tag{22}$$

It should be noted that the inertial frame in (20) is essentially $n(0)$. For initial alignment, $\mathbf{C}_n^i\mathbf{g}^n$ can be viewed as known input. The relationship between $\mathbf{r}^i$ and geographical $\mathbf{p}^n$ is given by

$$\mathbf{r}^i = \mathbf{C}_e^i\mathbf{r}^e = \mathbf{C}_e^i\ell\left(\mathbf{p}^n\right) \tag{23}$$

where

$$\mathbf{r}^e = \ell\left(\mathbf{p}^n\right) = \begin{bmatrix} \left(R_N + h\right)\cos L\cos\lambda \\ \left(R_N + h\right)\cos L\sin\lambda \\ \left[R_N\left(1 - e_1^2\right) + h\right]\sin L \end{bmatrix} \tag{24}$$

The attitude transformation matrix $\mathbf{C}_e^i$ is given by

$$\mathbf{C}_e^i = \mathbf{C}_{e(0)}^{n(0)}\mathbf{C}_e^{e(0)} \tag{25}$$

where $e(0)$ is the initial inertially fixed frame corresponding to earth frame $e$. Accordingly, $\mathbf{C}_e^{e(0)}$ encodes the attitude changes of the earth frame caused by the earth rotation and is given by

$$\mathbf{C}_{e(0)}^e = \begin{bmatrix} \cos\left(\omega_{ie}t\right) & \sin\left(\omega_{ie}t\right) & 0 \\ -\sin\left(\omega_{ie}t\right) & \cos\left(\omega_{ie}t\right) & 0 \\ 0 & 0 & 1 \end{bmatrix} \tag{26}$$

$\mathbf{C}_{e(0)}^{n(0)}$ is the function of the initial geographical position and is given by

$$\mathbf{C}_{e(0)}^{n(0)} = \begin{bmatrix} -\sin\lambda_0 & \cos\lambda_0 & 0 \\ -\sin L_0\cos\lambda_0 & -\sin L_0\sin\lambda_0 & \cos L_0 \\ \cos L_0\cos\lambda_0 & \cos L_0\sin\lambda_0 & \sin L_0 \end{bmatrix} \tag{27}$$

The ground velocity $\mathbf{v}^n$ can be derived as

$$\mathbf{v}^n = \mathbf{C}_e^n\mathbf{v}^e = \mathbf{C}_e^n\dot{\mathbf{r}}^e \tag{28}$$

The velocity $\mathbf{v}^i$ is given by

$$\mathbf{v}^i = \dot{\mathbf{r}}^i = \dot{\mathbf{C}}_e^i\mathbf{r}^e + \mathbf{C}_e^i\dot{\mathbf{r}}^e = \mathbf{C}_e^i\left(\boldsymbol{\omega}_{ie}^e \times\right)\mathbf{r}^e + \mathbf{C}_e^i\dot{\mathbf{r}}^e \tag{29}$$

Substituting (28) into (29) gives

$$\mathbf{v}^i = \mathbf{C}_e^i\left(\boldsymbol{\omega}_{ie}^e \times\right)\mathbf{r}^e + \mathbf{C}_n^i\mathbf{v}^n \tag{30}$$

This is the relationship between $\mathbf{v}^i$ and ground velocity $\mathbf{v}^n$. It can be used to initialize $\mathbf{v}^i$ for velocity calculation in (22) and can also be used as the measurement for the initial alignment.

## IV. SINS Error State Models based on $\mathbb{SO}(3)$

In traditional SINS error state equations, only the attitude is treated specially with consideration of its manifold structure. The reset of the state, including the velocity, position and inertial sensors' constant drift biases, are all treated as general vectors, that is $\boldsymbol{\delta}\mathbf{x} \in \mathbb{SO}(3) \times \mathbb{R}^n$ where $n$ is the dimension of the state other than attitude. In this paper, we term such error equations as error state equations based on $\mathbb{SO}(3)$, which is corresponding to the following error state equations based on $\mathbb{SE}_2(3)$. According to the expressed frames, there are two different attitude error definitions, one is reference frame attitude error and the other is body frame attitude error. For the decomposed model (22), the reference frame is the inertial frame. In this section, we will present the explicit error state equations for the two different attitude error definitions, which is the basis for the derivation of the $\mathbb{SE}_2(3)$ based error state equations in the next section.

In traditional SINS error model, the velocity and position are treated as vectors in Euclidean space and their errors are defined directly as

$$\boldsymbol{\delta}\mathbf{v}^i = \tilde{\mathbf{v}}^i - \mathbf{v}^i, \boldsymbol{\delta}\mathbf{r}^i = \tilde{\mathbf{r}}^i - \mathbf{r}^i \tag{31}$$

where $\tilde{\mathbf{v}}^i$ and $\tilde{\mathbf{r}}^i$ are the SINS calculated velocity and position making use of the formulation (22). There are two different attitude error definitions according to the expressed frames. The first one is the inertial frame attitude error, which is defined as

$$\boldsymbol{\delta}\mathbf{C} = \mathbf{C}_b^i \tilde{\mathbf{C}}_b^{iT} \tag{32}$$

In the above definition, the attitude error is difference between the inertial frame and calculated inertial frame, i.e.

$$\boldsymbol{\delta}\mathbf{C} = \mathbf{C}_b^i \tilde{\mathbf{C}}_b^{i'T} = \mathbf{C}_{i'}^i \tag{33}$$

With the small attitude error assumption, $\boldsymbol{\delta}\mathbf{C}$ can be approximated as

$$\mathbf{C}_{i'}^i = \mathbf{I}_{3\times3} + \left(\boldsymbol{\varphi}^i \times\right) \tag{34}$$

where $\boldsymbol{\varphi}^i$ is the attitude error in Euler angles form corresponding to $\boldsymbol{\delta}\mathbf{C}$ .With small attitude error assumption, the rotation sequence does not affect the result of the corresponding attitude matrix $\boldsymbol{\delta}\mathbf{C}$ . Actually, $\boldsymbol{\varphi}^i$ can also be viewed as a rotation vector, because that both the Euler angles and rotation vector are two times the vector part of the corresponding quaternion form [33].

Omitting the trivial derivation process, the error model corresponding to the inertial differential equations (22) in terms of $\boldsymbol{\varphi}^i$ , $\boldsymbol{\delta}\mathbf{v}^i$ and $\boldsymbol{\delta}\mathbf{r}^i$ is given by

$$\dot{\boldsymbol{\varphi}}^i = -\tilde{\mathbf{C}}_b^{i'} \boldsymbol{\delta\omega}_{ib}^b \tag{35a}$$

$$\boldsymbol{\delta}\dot{\mathbf{v}}^i = \left(\tilde{\mathbf{C}}_b^{i'} \tilde{\mathbf{f}}^b \times\right)\boldsymbol{\varphi}^i + \tilde{\mathbf{C}}_b^{i'} \boldsymbol{\delta}\mathbf{f}^b \tag{35b}$$

$$\boldsymbol{\delta}\dot{\mathbf{r}}^i = \boldsymbol{\delta}\mathbf{v}^i \tag{35c}$$

where $\boldsymbol{\delta\omega}_{ib}^b$ is the errors corresponding to $\boldsymbol{\omega}_{ib}^b$ ,and is defined as $\boldsymbol{\delta\omega}_{ib}^b = \tilde{\boldsymbol{\omega}}_{ib}^b - \boldsymbol{\omega}_{ib}^b$ . $\boldsymbol{\delta}\mathbf{f}^b$ is the errors corresponding to $\mathbf{f}^b$ ,and is defined as $\boldsymbol{\delta}\mathbf{f}^b = \tilde{\mathbf{f}}^b - \mathbf{f}^b$ .

For the inertial sensors errors, if we only consider the constant drift bias and noise, $\boldsymbol{\delta\omega}_{ib}^b$ and $\boldsymbol{\delta}\mathbf{f}^b$ can be expanded as

$$\boldsymbol{\delta\omega}_{ib}^b = \boldsymbol{\varepsilon}^b + \boldsymbol{\eta}_g^b \tag{36a}$$

$$\boldsymbol{\delta}\mathbf{f}^b = \nabla^b + \boldsymbol{\eta}_a^b \tag{36b}$$

where $\boldsymbol{\varepsilon}^b$ is gyroscope drift bias and $\boldsymbol{\eta}_g^b$ is the measurements noise of gyroscope. $\nabla^b$ is accelerometer drift bias and $\boldsymbol{\eta}_a^b$ is the measurements noise of accelerometer. If we augment the drift biases into the state vector, that is

$$\boldsymbol{\delta}\mathbf{x}_{\mathbf{Rso}} = \begin{bmatrix} \boldsymbol{\varphi}^{iT} & \boldsymbol{\delta}\mathbf{v}^{iT} & \boldsymbol{\delta}\mathbf{r}^{iT} & \boldsymbol{\varepsilon}^{bT} & \nabla^{bT} \end{bmatrix}^T \tag{37}$$

the corresponding state-space model is given by

$$\boldsymbol{\delta}\dot{\mathbf{x}}_{\mathbf{Rso}} = \mathbf{F}_{\mathbf{Rso}} \boldsymbol{\delta}\mathbf{x}_{\mathbf{Rso}} + \mathbf{G}_{\mathbf{Rso}} \begin{bmatrix} \boldsymbol{\eta}_g^b \\ \boldsymbol{\eta}_a^b \end{bmatrix} \tag{38}$$

where

$$\mathbf{F}_{\mathbf{Rso}} = \begin{bmatrix} \mathbf{0}_{3\times3} & \mathbf{0}_{3\times3} & \mathbf{0}_{3\times3} & -\tilde{\mathbf{C}}_b^{i'} & \mathbf{0}_{3\times3} \\ \left(\tilde{\mathbf{C}}_b^{i'} \tilde{\mathbf{f}}^b \times\right) & \mathbf{0}_{3\times3} & \mathbf{0}_{3\times3} & \mathbf{0}_{3\times3} & -\tilde{\mathbf{C}}_b^{i'} \\ \mathbf{0}_{3\times3} & \mathbf{I}_{3\times3} & \mathbf{0}_{3\times3} & \mathbf{0}_{3\times3} & \mathbf{0}_{3\times3} \\ \mathbf{0}_{3\times3} & \mathbf{0}_{3\times3} & \mathbf{0}_{3\times3} & \mathbf{0}_{3\times3} & \mathbf{0}_{3\times3} \\ \mathbf{0}_{3\times3} & \mathbf{0}_{3\times3} & \mathbf{0}_{3\times3} & \mathbf{0}_{3\times3} & \mathbf{0}_{3\times3} \end{bmatrix} \tag{39}$$

$$\mathbf{G}_{\mathbf{Rso}} = \begin{bmatrix} -\tilde{\mathbf{C}}_b^{i'} & \mathbf{0}_{3\times3} \\ \mathbf{0}_{3\times3} & -\tilde{\mathbf{C}}_b^{i'} \\ \mathbf{0}_{3\times3} & \mathbf{0}_{3\times3} \\ \mathbf{0}_{3\times3} & \mathbf{0}_{3\times3} \\ \mathbf{0}_{3\times3} & \mathbf{0}_{3\times3} \end{bmatrix} \tag{40}$$

The subscript **Rso** denotes the involved error state is based on $\mathbb{SO}(3)$ formulation with inertial frame (*right*) attitude error definition. It will be shown in the following context that the inertial frame attitude error is just corresponding to the right error definition in $\mathbb{SE}_2(3)$ .

Another attitude error form is the body frame attitude error, which is defined as

$$\boldsymbol{\delta}\mathbf{C} = \tilde{\mathbf{C}}_b^{iT} \mathbf{C}_b^i \tag{41}$$

In the above definition, the attitude error is the difference between the body frame and calculated body frame, i.e.

$$\boldsymbol{\delta}\mathbf{C} = \tilde{\mathbf{C}}_{b'}^{iT} \mathbf{C}_b^i = \mathbf{C}_b^{b'} \tag{42}$$

Similarly, with the small attitude error assumption, $\boldsymbol{\delta}\mathbf{C}$ can be approximated as

$$\mathbf{C}_b^{b'} = \mathbf{I}_{3\times3} + \left(\boldsymbol{\varphi}^b \times\right) \tag{43}$$

Similarly, we can also give the error model with such attitude error definition as [35]

$$\dot{\boldsymbol{\varphi}}^b = -\tilde{\boldsymbol{\omega}}_{ib}^b \times \boldsymbol{\varphi}^b - \boldsymbol{\delta\omega}_{ib}^b \tag{44a}$$

$$\boldsymbol{\delta}\dot{\mathbf{v}}^i = \tilde{\mathbf{C}}_{b'}^i \left(\tilde{\mathbf{f}}^b \times\right)\boldsymbol{\varphi}^b + \tilde{\mathbf{C}}_{b'}^i \boldsymbol{\delta}\mathbf{f}^b \tag{44b}$$

$$\boldsymbol{\delta}\dot{\mathbf{r}}^i = \boldsymbol{\delta}\mathbf{v}^i \tag{44c}$$

Similarly, define the augmented state vector as

$$\boldsymbol{\delta}\mathbf{x}_{\mathbf{Lso}} = \begin{bmatrix} \boldsymbol{\varphi}^{bT} & \boldsymbol{\delta}\mathbf{v}^{iT} & \boldsymbol{\delta}\mathbf{r}^{iT} & \boldsymbol{\varepsilon}^{bT} & \nabla^{bT} \end{bmatrix}^T \tag{45}$$

the state-space model for the left error definition is given by

$$\boldsymbol{\delta}\dot{\mathbf{x}}_{\mathbf{Lso}} = \mathbf{F}_{\mathbf{Lso}} \boldsymbol{\delta}\mathbf{x}_{\mathbf{Lso}} + \mathbf{G}_{\mathbf{Lso}} \begin{bmatrix} \boldsymbol{\eta}_g^b \\ \boldsymbol{\eta}_a^b \end{bmatrix} \tag{46}$$

where

$$\mathbf{F}_{\mathbf{Lso}} = \begin{bmatrix} -\left(\tilde{\boldsymbol{\omega}}_{ib}^b \times\right) & \mathbf{0}_{3\times3} & \mathbf{0}_{3\times3} & -\mathbf{I}_{3\times3} & \mathbf{0}_{3\times3} \\ \tilde{\mathbf{C}}_{b'}^i \left(\tilde{\mathbf{f}}^b \times\right) & \mathbf{0}_{3\times3} & \mathbf{0}_{3\times3} & \mathbf{0}_{3\times3} & \tilde{\mathbf{C}}_{b'}^i \\ \mathbf{0}_{3\times3} & \mathbf{I}_{3\times3} & \mathbf{0}_{3\times3} & \mathbf{0}_{3\times3} & \mathbf{0}_{3\times3} \\ \mathbf{0}_{3\times3} & \mathbf{0}_{3\times3} & \mathbf{0}_{3\times3} & \mathbf{0}_{3\times3} & \mathbf{0}_{3\times3} \\ \mathbf{0}_{3\times3} & \mathbf{0}_{3\times3} & \mathbf{0}_{3\times3} & \mathbf{0}_{3\times3} & \mathbf{0}_{3\times3} \end{bmatrix} \tag{47}$$

$$\mathbf{G}_{\mathbf{Lso}} = \begin{bmatrix} -\mathbf{I}_{3\times3} & \mathbf{0}_{3\times3} \\ \mathbf{0}_{3\times3} & \tilde{\mathbf{C}}_{b'}^i \\ \mathbf{0}_{3\times3} & \mathbf{0}_{3\times3} \\ \mathbf{0}_{3\times3} & \mathbf{0}_{3\times3} \\ \mathbf{0}_{3\times3} & \mathbf{0}_{3\times3} \end{bmatrix} \tag{48}$$

The subscript $\mathbf{Lso}$ denotes the involved error state is based on $\mathbb{SO}(3)$ formulation with body frame (*left*) attitude error definition. It will be shown in the following context that the body frame attitude error is just corresponding to the left error defined in $\mathbb{SE}_2(3)$.

It is shown in (44) that the velocity and position errors are still expressed in inertial frame. This is because that only the attitude has been specially treated and the other navigation parameters are all directly treated as vectors in Euclidean space. $\tilde{\mathbf{C}}_b^{i'}$ and $\tilde{\mathbf{C}}_{b'}^{i}$ are both obtained from SINS calculation, as will be shown in **Algorithm 1**. The difference of the marked frames is only used to emphasize the expressed frames of the defined attitude errors.

The $\mathbb{SO}(3)$ based error state models (35) and (44) are the basis for the derivation of the $\mathbb{SE}_2(3)$ based error state equations in the next section. It will be shown that, the inertial frame attitude error model (35) is corresponding to the right error model based on $\mathbb{SE}_2(3)$ and the body frame attitude error model (44) is corresponding to the left error model based on $\mathbb{SE}_2(3)$.

For initial alignment, the velocity provided by GNSS is always used as measurement. For the indirect filtering application, the filtering measurement is given by

$$\mathbf{y} = \tilde{\mathbf{v}}^i - \mathbf{v}^i = \boldsymbol{\delta}\mathbf{v}^i \tag{49}$$

$\mathbf{v}^i$ is given by (30) with known values provided by GNSS. The measurement transition equation corresponding to (49) is given by

$$\mathbf{y} = \underbrace{\begin{bmatrix} \mathbf{0}_{3\times3} & \mathbf{I}_{3\times3} & \mathbf{0}_{3\times9} \end{bmatrix}}_{\mathbf{H_{Rso}}/\mathbf{H_{Lso}}} \boldsymbol{\delta}\mathbf{x}_{\mathbf{Rso/Lso}} \tag{50}$$

Since the velocity error models in (35b) and (44b) are with the same definition form, the measurement transition equation (50) is suitable for both (37) and (45).

## V. SINS Damping Error State Models based on $\mathbb{SE}_2(3)$

In this section, we will derive the error state models based on $\mathbb{SE}_2(3)$. Both the right and left error definitions are considered. In contrast with the $\mathbb{SO}(3)$ based error models, the attitude matrix, velocity and position are formulated together as element of $\mathbb{SE}_2(3)$, that is

$$\mathbf{T} = \left[\begin{array}{c|cc} \mathbf{C}_b^i & \mathbf{v}^i & \mathbf{r}^i \\ \hline \mathbf{0}_{2\times3} & \multicolumn{2}{c}{\mathbf{I}_{2\times2}} \end{array}\right] \tag{51}$$

### A. Right Error State Model

Firstly, we consider the right error definition. The error-contaminated form of $\mathbf{T}$ is given by

$$\tilde{\mathbf{T}} = \left[\begin{array}{c|cc} \tilde{\mathbf{C}}_b^{i'} & \tilde{\mathbf{v}}^i & \tilde{\mathbf{r}}^i \\ \hline \mathbf{0}_{2\times3} & \multicolumn{2}{c}{\mathbf{I}_{2\times2}} \end{array}\right] \tag{52}$$

The corresponding right error is defined as

$$\begin{aligned} \boldsymbol{\delta}\mathbf{T}_{\mathbf{Rse}} &= \mathbf{T}\tilde{\mathbf{T}}^{-1} \\ &= \left[\begin{array}{c|cc} \mathbf{C}_b^i & \mathbf{v}^i & \mathbf{r}^i \\ \hline \mathbf{0}_{2\times3} & \multicolumn{2}{c}{\mathbf{I}_{2\times2}} \end{array}\right] \left[\begin{array}{c|cc} \tilde{\mathbf{C}}_b^{i'T} & -\tilde{\mathbf{C}}_b^{i'T}\tilde{\mathbf{v}}^i & -\tilde{\mathbf{C}}_b^{i'T}\tilde{\mathbf{r}}^i \\ \hline \mathbf{0}_{2\times3} & \multicolumn{2}{c}{\mathbf{I}_{2\times2}} \end{array}\right] \\ &= \left[\begin{array}{c|cc} \mathbf{C}_{i'}^i & \mathbf{v}^i - \mathbf{C}_{i'}^i\tilde{\mathbf{v}}^i & \mathbf{r}^i - \mathbf{C}_{i'}^i\tilde{\mathbf{r}}^i \\ \hline \mathbf{0}_{2\times3} & \multicolumn{2}{c}{\mathbf{I}_{2\times2}} \end{array}\right] \end{aligned} \tag{53}$$

The subscript $\mathbf{Rse}$ denotes the involved error state is based on $\mathbb{SE}_2(3)$ formulation with *right* error definition. In second line of (53), we have made use of (2). Since the filtering is designed in an indirect manner, we would like to derive the vector form corresponding to $\boldsymbol{\delta}\mathbf{T}_{\mathbf{Rse}}$, i.e.

$$\mathbf{dx} = \ln\left(\boldsymbol{\delta}\mathbf{T}_{\mathbf{Rse}}\right)^{\vee} \tag{54}$$

Denote $\mathbf{dx} = \begin{bmatrix} \boldsymbol{\varphi}^{iT} & \mathbf{dv}^{iT} & \mathbf{dr}^{iT} \end{bmatrix}^T$. $\boldsymbol{\varphi}^i$ is corresponding to $\mathbf{C}_{i'}^i$ and its model can be readily obtained from (35a). According to (10) we have

$$\mathbf{dv}^i = \mathbf{J}_l^{-1}\left(\mathbf{v}^i - \mathbf{C}_{i'}^i\tilde{\mathbf{v}}^i\right) \tag{55}$$

$$\mathbf{dr}^i = \mathbf{J}_l^{-1}\left(\mathbf{r}^i - \mathbf{C}_{i'}^i\tilde{\mathbf{r}}^i\right) \tag{56}$$

In this paper, we aim to derive the linear error model. In this respect, the attitude error is assumed to be small. With such assumption, $\boldsymbol{\varphi}^i$ can be derived according to (34). Meanwhile, $\mathbf{J}_l^{-1}$ can be approximated as

$$\mathbf{J}_l^{-1} = \mathbf{I}_{3\times3} - \frac{1}{2}\left(\boldsymbol{\varphi}^e \times\right) \tag{57}$$

Substituting (34) and (57) into (55) and (56) gives

$$\mathbf{dv}^i = -\boldsymbol{\delta}\mathbf{v}^i + \tilde{\mathbf{v}}^i \times \boldsymbol{\varphi}^i \tag{58}$$

$$\mathbf{dr}^i = -\boldsymbol{\delta}\mathbf{r}^i + \tilde{\mathbf{r}}^i \times \boldsymbol{\varphi}^i \tag{59}$$

It is shown that the velocity error in (58) is similar with the transformed velocity error in [36-38]. Actually, the transformed velocity error can be readily obtained through formulating the attitude and velocity as element of $\mathbb{SE}(3)$. Actually, the geometric gyroscope drift bias error in [39-41] can also be derived in a similar manner as pointed out in [42].

The differential equation of $\mathbf{dv}^i$ is given by

$$\begin{aligned} \mathbf{d}\dot{\mathbf{v}}^i &= -\boldsymbol{\delta}\dot{\mathbf{v}}^i + \dot{\tilde{\mathbf{v}}}^i \times \boldsymbol{\varphi}^i + \tilde{\mathbf{v}}^i \times \dot{\boldsymbol{\varphi}}^i \\ &= -\tilde{\mathbf{C}}_b^{i'}\tilde{\mathbf{f}}^b \times \boldsymbol{\varphi}^i - \tilde{\mathbf{C}}_b^{i'}\boldsymbol{\delta}\mathbf{f}^b + \left(\tilde{\mathbf{C}}_b^{i'}\tilde{\mathbf{f}}^b + \mathbf{g}^i\right) \times \boldsymbol{\varphi}^i - \tilde{\mathbf{v}}^i \times \left(\tilde{\mathbf{C}}_b^{i'}\boldsymbol{\delta\omega}_{ib}^b\right) \\ &= \left(\mathbf{g}^i \times\right)\boldsymbol{\varphi}^i - \left(\tilde{\mathbf{v}}^i \times\right)\tilde{\mathbf{C}}_b^{i'}\boldsymbol{\delta\omega}_{ib}^b - \tilde{\mathbf{C}}_b^{i'}\boldsymbol{\delta}\mathbf{f}^b \end{aligned} \tag{60}$$

where $\mathbf{g}^i = \mathbf{C}_n^i\mathbf{g}^n$.

The differential equation of $\mathbf{dr}^i$ is given by

$$\begin{aligned} \mathbf{d}\dot{\mathbf{r}}^i &= -\boldsymbol{\delta}\mathbf{r}^i + \dot{\tilde{\mathbf{r}}}^i \times \boldsymbol{\varphi}^i + \tilde{\mathbf{r}}^i \times \dot{\boldsymbol{\varphi}}^i \\ &= -\boldsymbol{\delta}\mathbf{v}^i + \tilde{\mathbf{v}}^i \times \boldsymbol{\varphi}^i - \left(\tilde{\mathbf{r}}^i \times\right)\left(\tilde{\mathbf{C}}_b^{i'}\boldsymbol{\delta\omega}_{ib}^b\right) \\ &= \mathbf{dv}^i - \left(\tilde{\mathbf{r}}^i \times\right)\tilde{\mathbf{C}}_b^{i'}\boldsymbol{\delta\omega}_{ib}^b \end{aligned} \tag{61}$$

If we augment the drift biases into the state vector, that is

$$\mathbf{dx}_{\mathbf{Rse}} = \begin{bmatrix} \boldsymbol{\varphi}^{iT} & \mathbf{dv}^{iT} & \mathbf{dr}^{iT} & \boldsymbol{\varepsilon}^{bT} & \nabla^{bT} \end{bmatrix}^T \tag{62}$$

the corresponding state-space model is given by

$$\mathbf{d}\dot{\mathbf{x}}_{\mathbf{Rse}}=\mathbf{F}_{\mathbf{Rse}}\mathbf{dx}_{\mathbf{Rse}}+\mathbf{G}_{\mathbf{Rse}}\begin{bmatrix}\boldsymbol{\eta}_g^b\\ \boldsymbol{\eta}_a^b\end{bmatrix} \tag{63}$$

where

$$\mathbf{F}_{\mathbf{Rse}}=\begin{bmatrix}\mathbf{0}_{3\times3} & \mathbf{0}_{3\times3} & \mathbf{0}_{3\times3} & -\tilde{\mathbf{C}}_b^{i'} & \mathbf{0}_{3\times3}\\ (\mathbf{g}^i\times) & \mathbf{0}_{3\times3} & \mathbf{0}_{3\times3} & -(\tilde{\mathbf{v}}^i\times)\tilde{\mathbf{C}}_b^{i'} & -\tilde{\mathbf{C}}_b^{i'}\\ \mathbf{0}_{3\times3} & \mathbf{I}_{3\times3} & \mathbf{0}_{3\times3} & -(\tilde{\mathbf{r}}^i\times)\tilde{\mathbf{C}}_b^{i'} & \mathbf{0}_{3\times3}\\ \mathbf{0}_{3\times3} & \mathbf{0}_{3\times3} & \mathbf{0}_{3\times3} & \mathbf{0}_{3\times3} & \mathbf{0}_{3\times3}\\ \mathbf{0}_{3\times3} & \mathbf{0}_{3\times3} & \mathbf{0}_{3\times3} & \mathbf{0}_{3\times3} & \mathbf{0}_{3\times3}\end{bmatrix} \tag{64}$$

$$\mathbf{G}_{\mathbf{Rse}}=\begin{bmatrix}-\tilde{\mathbf{C}}_b^{i'} & \mathbf{0}_{3\times3}\\ -(\tilde{\mathbf{v}}^i\times)\tilde{\mathbf{C}}_b^{i'} & -\tilde{\mathbf{C}}_b^{i'}\\ -(\tilde{\mathbf{r}}^i\times)\tilde{\mathbf{C}}_b^{i'} & \mathbf{0}_{3\times3}\\ \mathbf{0}_{3\times3} & \mathbf{0}_{3\times3}\\ \mathbf{0}_{3\times3} & \mathbf{0}_{3\times3}\end{bmatrix} \tag{65}$$

The velocity error measurement equation is given by

$$\mathbf{y}=\tilde{\mathbf{v}}^i-\mathbf{v}^i=\boldsymbol{\delta}\mathbf{v}^i=\tilde{\mathbf{v}}^i\times\boldsymbol{\varphi}^i-\mathbf{dv}^i \tag{66}$$

The measurement transition equation corresponding to (66) is given by

$$\mathbf{y}=\underbrace{\begin{bmatrix}(\tilde{\mathbf{v}}^i\times) & -\mathbf{I}_{3\times3} & \mathbf{0}_{3\times9}\end{bmatrix}}_{\mathbf{H}_{\mathbf{Rse}}}\mathbf{dx}_{\mathbf{Rse}} \tag{67}$$

## B. *Left Error State Model*

Denote the error-contaminated form of $\mathbf{T}$ for left error definition as

$$\tilde{\mathbf{T}}=\left[\begin{array}{c|cc}\tilde{\mathbf{C}}_{b'}^i & \tilde{\mathbf{v}}^i & \tilde{\mathbf{r}}^i\\ \hline \mathbf{0}_{2\times3} & \multicolumn{2}{c}{\mathbf{I}_{2\times2}}\end{array}\right] \tag{68}$$

The left error is given by

$$\begin{aligned}\boldsymbol{\delta}\mathbf{T}_{\mathbf{Lse}}&=\tilde{\mathbf{T}}^{-1}\mathbf{T}\\ &=\left[\begin{array}{c|cc}\tilde{\mathbf{C}}_{b'}^{iT} & -\tilde{\mathbf{C}}_{b'}^{iT}\tilde{\mathbf{v}}^i & -\tilde{\mathbf{C}}_{b'}^{iT}\tilde{\mathbf{r}}^i\\ \hline \mathbf{0}_{2\times3} & \multicolumn{2}{c}{\mathbf{I}_{2\times2}}\end{array}\right]\left[\begin{array}{c|cc}\mathbf{C}_b^i & \mathbf{v}^i & \mathbf{r}^i\\ \hline \mathbf{0}_{2\times3} & \multicolumn{2}{c}{\mathbf{I}_{2\times2}}\end{array}\right]\\ &=\left[\begin{array}{c|cc}\mathbf{C}_b^{b'} & \tilde{\mathbf{C}}_{b'}^{i\ T}(\mathbf{v}^i-\tilde{\mathbf{v}}^i) & \tilde{\mathbf{C}}_{b'}^{i\ T}(\mathbf{r}^i-\tilde{\mathbf{r}}^i)\\ \hline \mathbf{0}_{2\times3} & \multicolumn{2}{c}{\mathbf{I}_{2\times2}}\end{array}\right]\end{aligned} \tag{69}$$

The subscript **Lse** denotes the involved error state is based on $\mathbb{SE}_2(3)$ formulation with *left* error definition.

Denote $\mathbf{dx}=\begin{bmatrix}\boldsymbol{\varphi}^{bT} & \mathbf{dv}^{bT} & \mathbf{dr}^{bT}\end{bmatrix}^T$ as the vector form corresponding to $\boldsymbol{\delta}\mathbf{T}_{\mathbf{Lse}}$ in (69). $\boldsymbol{\varphi}^b$ is corresponding to $\mathbf{C}_b^{b'}$ and its model can be readily obtained from (44a). According to (10) we have

$$\mathbf{dv}^b=\mathbf{J}_l^{-1}\tilde{\mathbf{C}}_{b'}^{i\ T}\left(\mathbf{v}^i-\tilde{\mathbf{v}}^i\right) \tag{70}$$

$$\mathbf{dr}^b=\mathbf{J}_l^{-1}\tilde{\mathbf{C}}_{b'}^{i\ T}\left(\mathbf{r}^i-\tilde{\mathbf{r}}^i\right) \tag{71}$$

With the similar assumption in (57), (70) and (71) can be approximated as

$$\mathbf{dv}^b\approx\tilde{\mathbf{C}}_{b'}^{i\ T}\left(\mathbf{v}^i-\tilde{\mathbf{v}}^i\right)=-\tilde{\mathbf{C}}_{b'}^{i\ T}\boldsymbol{\delta}\mathbf{v}^i \tag{72}$$

$$\mathbf{dr}^b\approx\tilde{\mathbf{C}}_{b'}^{i\ T}\left(\mathbf{r}^i-\tilde{\mathbf{r}}^i\right)=-\tilde{\mathbf{C}}_{b'}^{i\ T}\boldsymbol{\delta}\mathbf{r}^i \tag{73}$$

The differential equation of $\mathbf{dv}^b$ in (72) is given by

$$\begin{aligned}\mathbf{d}\dot{\mathbf{v}}^b&=-\dot{\tilde{\mathbf{C}}}_{b'}^{i\ T}\boldsymbol{\delta}\mathbf{v}^i-\tilde{\mathbf{C}}_{b'}^{i\ T}\boldsymbol{\delta}\dot{\mathbf{v}}^i\\ &=\left(\tilde{\boldsymbol{\omega}}_{ib}^b\times\right)\tilde{\mathbf{C}}_{b'}^{i\ T}\boldsymbol{\delta}\mathbf{v}^i-\tilde{\mathbf{C}}_{b'}^{i\ T}\left[\tilde{\mathbf{C}}_{b'}^i\left(\tilde{\mathbf{f}}^b\times\right)\boldsymbol{\varphi}^b+\tilde{\mathbf{C}}_{b'}^i\boldsymbol{\delta}\mathbf{f}^b\right]\\ &=-\left(\tilde{\mathbf{f}}^b\times\right)\boldsymbol{\varphi}^b-\left(\tilde{\boldsymbol{\omega}}_{ib}^b\times\right)\mathbf{dv}^i-\boldsymbol{\delta}\mathbf{f}^b\end{aligned} \tag{74}$$

The differential equation of $\mathbf{dr}^b$ in (73) is given by

$$\begin{aligned}\mathbf{d}\dot{\mathbf{r}}^b&=-\dot{\tilde{\mathbf{C}}}_{b'}^{i\ T}\boldsymbol{\delta}\mathbf{r}^i-\tilde{\mathbf{C}}_{b'}^{i\ T}\boldsymbol{\delta}\dot{\mathbf{r}}^i\\ &=\mathbf{dv}^b-\left(\tilde{\boldsymbol{\omega}}_{ib}^b\times\right)\mathbf{dr}^b\end{aligned} \tag{75}$$

Similarly, define the augmented state vector as

$$\mathbf{dx}_{\mathbf{Lse}}=\begin{bmatrix}\boldsymbol{\varphi}^{bT} & \mathbf{dv}^{bT} & \mathbf{dr}^{bT} & \boldsymbol{\varepsilon}^{bT} & \nabla^{bT}\end{bmatrix}^T \tag{76}$$

the state-space model for the left error definition is given by

$$\mathbf{d}\dot{\mathbf{x}}_{\mathbf{Lse}}=\mathbf{F}_{\mathbf{Lse}}\mathbf{dx}_{\mathbf{Lse}}+\mathbf{G}_{\mathbf{Lse}}\begin{bmatrix}\boldsymbol{\eta}_g^b\\ \boldsymbol{\eta}_a^b\end{bmatrix} \tag{77}$$

where

$$\mathbf{F}_{\mathbf{Lse}}=\begin{bmatrix}-\left(\tilde{\boldsymbol{\omega}}_{ib}^b\times\right) & \mathbf{0}_{3\times3} & \mathbf{0}_{3\times3} & -\mathbf{I}_{3\times3} & \mathbf{0}_{3\times3}\\ -\left(\tilde{\mathbf{f}}^b\times\right) & -\left(\tilde{\boldsymbol{\omega}}_{ib}^b\times\right) & \mathbf{0}_{3\times3} & \mathbf{0}_{3\times3} & -\mathbf{I}_{3\times3}\\ \mathbf{0}_{3\times3} & \mathbf{I}_{3\times3} & -\left(\tilde{\boldsymbol{\omega}}_{ib}^b\times\right) & \mathbf{0}_{3\times3} & \mathbf{0}_{3\times3}\\ \mathbf{0}_{3\times3} & \mathbf{0}_{3\times3} & \mathbf{0}_{3\times3} & \mathbf{0}_{3\times3} & \mathbf{0}_{3\times3}\\ \mathbf{0}_{3\times3} & \mathbf{0}_{3\times3} & \mathbf{0}_{3\times3} & \mathbf{0}_{3\times3} & \mathbf{0}_{3\times3}\end{bmatrix} \tag{78}$$

$$\mathbf{G}_{\mathbf{Lse}}=\begin{bmatrix}-\mathbf{I}_{3\times3} & \mathbf{0}_{3\times3}\\ \mathbf{0}_{3\times3} & -\mathbf{I}_{3\times3}\\ \mathbf{0}_{3\times3} & \mathbf{0}_{3\times3}\\ \mathbf{0}_{3\times3} & \mathbf{0}_{3\times3}\\ \mathbf{0}_{3\times3} & \mathbf{0}_{3\times3}\end{bmatrix} \tag{79}$$

The velocity error measurement equation is given by

$$\mathbf{y}=\tilde{\mathbf{v}}^i-\mathbf{v}^i=\boldsymbol{\delta}\mathbf{v}^i=-\tilde{\mathbf{C}}_{b'}^i\mathbf{dv}^b \tag{80}$$

The measurement transition equation corresponding to (80) is given by

$$\mathbf{y}=\underbrace{\begin{bmatrix}\mathbf{0}_{3\times3} & -\tilde{\mathbf{C}}_{b'}^i & \mathbf{0}_{3\times9}\end{bmatrix}}_{\mathbf{H}_{\mathbf{Lse}}}\mathbf{dx}_{\mathbf{Lse}} \tag{81}$$

# VI. SINS Initial Alignment Procedures Based on $\mathbb{SE}_2(3)$ Error State Models

The explicit indirect SINS initial alignment procedures at time instant $k$ are presented in **Algorithm 1**.

---

**Algorithm1**: Indirect Initial Alignment based on KF

---

***1.****Perform SINS calculation using inertial differential equations (22)*

$$\left[\tilde{\mathbf{C}}_{b,k}^i,\tilde{\mathbf{v}}_k^i,\tilde{\mathbf{r}}_k^i\right]=\mathrm{SINS}\left[\hat{\mathbf{C}}_{b,k-1}^i,\hat{\mathbf{v}}_{k-1}^i,\hat{\mathbf{r}}_{k-1}^i,\tilde{\boldsymbol{\omega}}_{ib,k}^b,\tilde{\mathbf{f}}_k^b\right]$$

***2.*** *Construct the error state space model as in (38), (46) and (63), (77).*

***3.*** *Using the linear KF to estimate the error state*

$$\left[\mathbf{d}\hat{\mathbf{x}}_k,\mathbf{P}_k\right]=\mathrm{KF}\left[\mathbf{d}\hat{\mathbf{x}}_{k-1},\mathbf{P}_{k-1},\mathbf{F}_{k-1},\mathbf{H}_k,\mathbf{y}_k,\mathbf{Q}_{k-1},\mathbf{R}_k\right]$$

***4.*** *Refine the SINS calculated results using the estimated error state*

$$\left[\hat{\mathbf{C}}_{b,k}^i,\hat{\mathbf{v}}_k^i,\hat{\mathbf{r}}_k^i\right]=\left[\tilde{\mathbf{C}}_{b,k}^i,\tilde{\mathbf{v}}_k^i,\tilde{\mathbf{r}}_k^i\right]\boxplus\mathbf{d}\hat{\mathbf{x}}_k(1:9)$$

***5.*** *Reset the error state estimate*

---

$\mathbf{d}\hat{\mathbf{x}}_k(1:9)=\mathbf{0}_{9\times1}$

***6.** Go to the next time recursion.*

In **Algorithm 1**, the function $\text{SINS}[\cdots]$ is the SINS calculation algorithm corresponding to the inertial differential equations (22). One can apply the single-sample, two-sample or more advanced calculation algorithms to solve the differential equations [43]. For the SINS calculation, the known geographical velocity and position is used to calculate $\mathbf{C}_n^i$ and to initialize $\mathbf{v}^i$ and $\mathbf{r}^i$ according to (30) and (23). The calculated velocity $\mathbf{v}^i$ is also used as measurement for KF.

The function $\text{KF}[\cdots]$ is the general KF algorithm. $\mathbf{F}_{k-1}$ and $\mathbf{H}_k$ are the discrete-time forms of the transition matrices in (39), (47), (63), (78) and (50), (67), (81). $\mathbf{Q}_{k-1}$ is the process noise covariance and $\mathbf{R}_k$ is the measurement noise covariance. $\mathbf{y}_k=\tilde{\mathbf{v}}_k^i-\mathbf{v}_{\text{GPS},k}^i$ is the measurement.

The retraction operation $\boxplus$ denotes the modification process of the navigation parameters by the filtering estimate (in error form). Since the navigation parameters have been constructed as element of $\mathbb{SE}_2(3)$, the state estimate $\mathbf{d}\hat{\mathbf{x}}_k(1:9)=\begin{bmatrix}\hat{\boldsymbol{\varphi}}_k^T & \mathbf{d}\hat{\mathbf{v}}_k^T & \mathbf{d}\hat{\mathbf{r}}_k^T\end{bmatrix}^T$ should be firstly mapped into its $\mathbb{SE}_2(3)$ form. This can be done according to (6), i.e.

$$\boldsymbol{\delta}\hat{\mathbf{T}}_k=\left[\begin{array}{c|cc}\exp(\hat{\boldsymbol{\varphi}}_k) & \mathbf{J}_l\mathbf{d}\hat{\mathbf{v}}_k & \mathbf{J}_l\mathbf{d}\hat{\mathbf{r}}_k \\ \hline \mathbf{0}_{2\times3} & \multicolumn{2}{c}{\mathbf{I}_{2\times2}}\end{array}\right] \quad (82)$$

For the right error definition (53), the refined navigation parameters are given by

$$\begin{aligned}\hat{\mathbf{T}}_k&=\boldsymbol{\delta}\hat{\mathbf{T}}_{\mathbf{Rse},k}\tilde{\mathbf{T}}_k\\ &=\left[\begin{array}{c|cc}\exp(\hat{\boldsymbol{\varphi}}_k^i) & \mathbf{J}_l\mathbf{d}\hat{\mathbf{v}}_k^i & \mathbf{J}_l\mathbf{d}\hat{\mathbf{r}}_k^i \\ \hline \mathbf{0}_{2\times3} & \multicolumn{2}{c}{\mathbf{I}_{2\times2}}\end{array}\right]\left[\begin{array}{c|cc}\tilde{\mathbf{C}}_{b,k}^i & \tilde{\mathbf{v}}_k^i & \tilde{\mathbf{r}}_k^i \\ \hline \mathbf{0}_{2\times3} & \multicolumn{2}{c}{\mathbf{I}_{2\times2}}\end{array}\right]\\ &=\left[\begin{array}{c|cc}\exp(\hat{\boldsymbol{\varphi}}_k^i)\tilde{\mathbf{C}}_{b,k}^i & \exp(\hat{\boldsymbol{\varphi}}_k^i)\tilde{\mathbf{v}}_k^i+\mathbf{J}_l\mathbf{d}\hat{\mathbf{v}}_k^i & \exp(\hat{\boldsymbol{\varphi}}_k^i)\tilde{\mathbf{r}}_k^i+\mathbf{J}_l\mathbf{d}\hat{\mathbf{r}}_k^i \\ \hline \mathbf{0}_{2\times3} & \multicolumn{2}{c}{\mathbf{I}_{2\times2}}\end{array}\right]\end{aligned} \quad (83)$$

The modified navigation parameters can be readily obtained from (83) as

$$\hat{\mathbf{C}}_{b,k}^i=\exp(\hat{\boldsymbol{\varphi}}_k^i)\tilde{\mathbf{C}}_{b,k}^i \quad (84a)$$

$$\hat{\mathbf{v}}_k^i=\exp(\hat{\boldsymbol{\varphi}}_k^i)\tilde{\mathbf{v}}_k^i+\mathbf{J}_l\mathbf{d}\hat{\mathbf{v}}_k^i \quad (84b)$$

$$\hat{\mathbf{r}}_k^i=\exp(\hat{\boldsymbol{\varphi}}_k^i)\tilde{\mathbf{r}}_k^i+\mathbf{J}_l\mathbf{d}\hat{\mathbf{r}}_k^i \quad (84c)$$

Eqs. (84) are the closed-form modifications of the navigation parameters based on the error estimate. Based on the small attitude error assumption, which has been used for derivation of the error state model, (84b) and (84c) can be further approximated as

$$\hat{\mathbf{v}}_k^i=\tilde{\mathbf{v}}_k^i+\mathbf{d}\hat{\mathbf{v}}_k^i-\tilde{\mathbf{v}}_k^i\times\hat{\boldsymbol{\varphi}}_k^i \quad (85a)$$

$$\hat{\mathbf{r}}_k^i=\tilde{\mathbf{r}}_k^i+\mathbf{d}\hat{\mathbf{r}}_k^i-\tilde{\mathbf{r}}_k^i\times\hat{\boldsymbol{\varphi}}_k^i \quad (85b)$$

One can check that (85a) and (85b) are just corresponding to (58) and (59).

Similarly, for the left error definition (69), the refined navigation parameters are given by

$$\begin{aligned}\hat{\mathbf{T}}_k&=\tilde{\mathbf{T}}_k\boldsymbol{\delta}\hat{\mathbf{T}}_{\mathbf{Lse},k}\\ &=\left[\begin{array}{c|cc}\tilde{\mathbf{C}}_{b,k}^i & \tilde{\mathbf{v}}_k^i & \tilde{\mathbf{r}}_k^i \\ \hline \mathbf{0}_{2\times3} & \multicolumn{2}{c}{\mathbf{I}_{2\times2}}\end{array}\right]\left[\begin{array}{c|cc}\exp(\hat{\boldsymbol{\varphi}}_k^b) & \mathbf{J}_l\mathbf{d}\hat{\mathbf{v}}_k^b & \mathbf{J}_l\mathbf{d}\hat{\mathbf{r}}_k^b \\ \hline \mathbf{0}_{2\times3} & \multicolumn{2}{c}{\mathbf{I}_{2\times2}}\end{array}\right]\\ &=\left[\begin{array}{c|cc}\tilde{\mathbf{C}}_{b,k}^i\exp(\hat{\boldsymbol{\varphi}}_k^b) & \tilde{\mathbf{C}}_{b,k}^i\mathbf{J}_l\mathbf{d}\hat{\mathbf{v}}_k^b+\tilde{\mathbf{v}}_k^i & \tilde{\mathbf{C}}_{b,k}^i\mathbf{J}_l\mathbf{d}\hat{\mathbf{r}}_k^b+\tilde{\mathbf{r}}_k^i \\ \hline \mathbf{0}_{2\times3} & \multicolumn{2}{c}{\mathbf{I}_{2\times2}}\end{array}\right]\end{aligned} \quad (86)$$

The modified navigation parameters can be readily obtained from (86) as

$$\hat{\mathbf{C}}_{b,k}^i=\tilde{\mathbf{C}}_{b,k}^i\exp(\hat{\boldsymbol{\varphi}}_k^b) \quad (87a)$$

$$\hat{\mathbf{v}}_k^i=\tilde{\mathbf{C}}_{b,k}^i\mathbf{J}_l\mathbf{d}\hat{\mathbf{v}}_k^b+\tilde{\mathbf{v}}_k^i \quad (87b)$$

$$\hat{\mathbf{r}}_k^i=\tilde{\mathbf{C}}_{b,k}^i\mathbf{J}_l\mathbf{d}\hat{\mathbf{r}}_k^b+\tilde{\mathbf{r}}_k^i \quad (87c)$$

Similarly, (87b) and (87c) can also be further approximated as

$$\hat{\mathbf{v}}_k^i=\tilde{\mathbf{C}}_{b,k}^i\mathbf{d}\hat{\mathbf{v}}_k^b+\tilde{\mathbf{v}}_k^i \quad (88a)$$

$$\hat{\mathbf{r}}_k^i=\tilde{\mathbf{C}}_{b,k}^i\mathbf{d}\hat{\mathbf{r}}_k^b+\tilde{\mathbf{r}}_k^i \quad (88b)$$

which are just corresponding to (72) and (73).

**Algorithm 1** is mainly used to derive the attitude estimate $\hat{\mathbf{C}}_{b,k}^i$. The attitude estimate $\hat{\mathbf{C}}_{b,k}^n$ can be readily obtained through (18). The initial alignment procedure with the decomposed model is shown in Fig. 1. In this paper, four different linear state-space models are investigated. It will be discussed in next section that only the left error state model based on $\mathbb{SE}_2(3)$ is independent on the SINS calculated results and this is just the rooted reason for its superior performance.

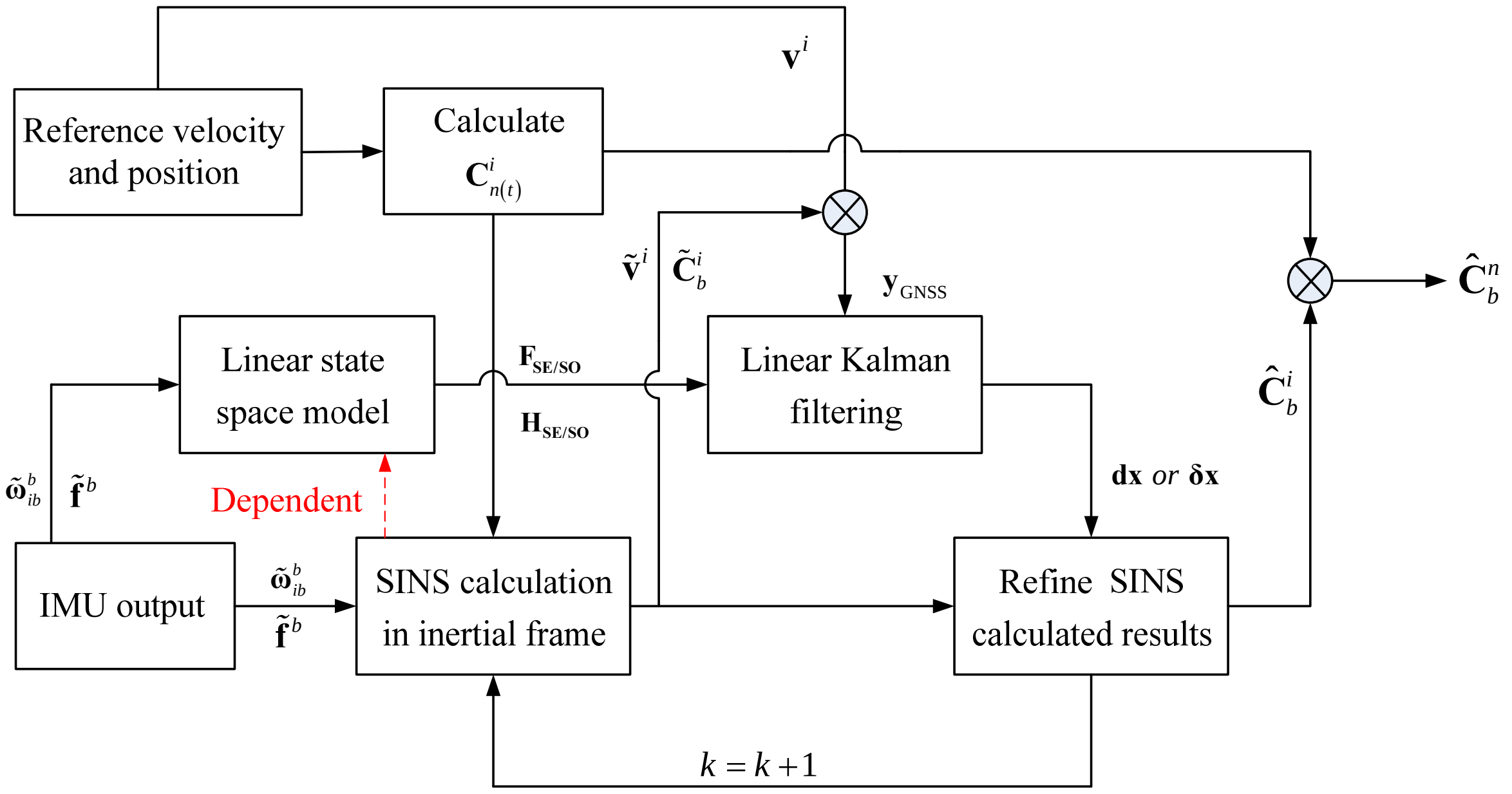

Fig. 1. Diagram for the initial alignment method based on attitude decomposition

It should be noted that the right and left state error definitions are seemingly a little different with those in [22-28]. This is because that the definitions in this work are consistent with the traditional attitude error definitions as those in section IV for indirect initial alignment applications. However, the definitions in the presented work are essentially also consistent with those in [22-28]. Take the right attitude error for example, the attitude errors defined in this work and that in [22-28] are both formulated to represent the errors expressed in navigation frame. The only difference is that the attitude error in this work represents the attitude transformation from true navigation frame to calculated navigation frame, while that in [22-28] represent the attitude transformation from estimated navigation frame to true navigation frame. That is to say, there is a transposition relationship between the two definitions. Such difference does not affect the performance of the resulted error model.

## VII. Validity Analysis of $\mathcal{SE}_2(3)$ Based Model for Linear Initial Alignment with Arbitrary Misalignments

In this section, the validity analysis of $\mathcal{SE}_2(3)$ based model for linear initial alignment with arbitrary misalignments is presented.

With the group state definition in (51), the SINS model in inertial frame (22) can be rewritten as function of the group state as

$$\Gamma_{\mathbf{u}}(\mathbf{T}) = \begin{bmatrix} \mathbf{C}^i_b \boldsymbol{\omega}^b_{ib} \times & \mathbf{C}^i_b \mathbf{f}^b + \mathbf{C}^i_n \mathbf{g}^n & \mathbf{v}^i \\ 0_{1\times 3} & 0 & 0 \\ 0_{1\times 3} & 0 & 0 \end{bmatrix} \tag{89}$$

In the following, we will demonstrate that the dynamic model (89) is a *group affine* system and its error model has an interesting *log-linear* property. Such striking property is just the fundamental of linear KF based initial alignment with arbitrary large misalignment.

**Theorem 1**: The dynamics model (89) satisfy the relationship

$$\Gamma_{\mathbf{u}}(\mathbf{T}_1\mathbf{T}_2) = \Gamma_{\mathbf{u}}(\mathbf{T}_1)\mathbf{T}_2 + \mathbf{T}_1\Gamma_{\mathbf{u}}(\mathbf{T}_2) - \mathbf{T}_1\Gamma_{\mathbf{u}}(\mathbf{I}_{5\times 5})\mathbf{T}_2 \tag{90}$$

where $\mathbf{T}_1, \mathbf{T}_2 \in \mathcal{SE}_2(3)$ are the realizations of the state (50). $\mathbf{u}$ is the input of the model (88), which include $\boldsymbol{\omega}^b_{ib}$, $\mathbf{f}^b$ and $\mathbf{C}^i_n \mathbf{g}^n$.

The explicit proof of **Theorem 1** is presented in Appendix. A dynamic model satisfying (90) is called group affine system. An attractive property of the group affine system is that its error dynamics are trajectory independent. As can be seen from (64) and (78) that, the error state models (only consider the navigation parameters) are indeed independent on the global state estimate. Such attractive property on one hand can remedy the inconsistency problem involved in inertial based applications and on the other hand, can make the vector error of the system satisfy a “log-linear” autonomous differential equation. Moreover, it is pointed out in [28] that the nonlinear group state error can be exactly recovered from such “log-linear” error state vector. This is just the theoretical basis for the linear initial alignment with arbitrary large misalignments.

***Remark 1:*** In the aforementioned discussion, we have not considered the inertial sensors’ errors. Unfortunately, if the inertial sensors’ errors are incorporated into the group state, the resulting model will not satisfy the group affine property. However, during the initial alignment, it is always desired to estimate the drift biases of the inertial sensors coupled with the attitude. Although under static or swaying conditions, some elements of the drift biases will not be observed, multi-position alignment procedures can be applied to remedy such problem. Fortunately, as pointed out in [28, 45], much of the advantages caused by the group affine property can be maintained in the augmented models.

***Remark 2:*** In order to estimate the nonlinear group state making use of the linear vector state model, another prerequisite is that the observation should also be with the *invariant* form.

The invariant observations are defined as [28, 45]

$$\text{Left-Invariant Observation: } \mathbf{y} = \boldsymbol{\chi}\mathbf{b} \tag{91a}$$

$$\text{Rright-Invariant Observation: } \mathbf{y} = \boldsymbol{\chi}^{-1}\mathbf{b} \tag{91b}$$

where $\mathbf{b}$ is a constant vector. With the invariant observation, the corresponding measurement transition matrix will also be independent on the global state estimate.

According to (91), it can be verified that the velocity $\mathbf{v}^i_{\text{GPS}}$ is a left-invariant observation. In this respect, the left error state definition will be more preferred. Moreover, with the left error definition, the corresponding transition matrix will be independent on the global state estimate. However, as can be seen from (81) that the measurement transition matrix $\mathbf{H}_{\mathbf{Lse}}$ is the function of $\tilde{\mathbf{C}}^i_{b'}$. It is seemingly that $\mathbf{H}_{\mathbf{Lse}}$ is not independent on the global state estimate. Fortunately, [26] has pointed out that "*Applying a linear function to the innovation term of an EKF before computing the gains does not change the results of the filter*". According to such statement, the coefficient $\tilde{\mathbf{C}}^i_{b'}$ in (81) can be moved to the innovation and therefore making the measurement equation be trajectory independent. That is to say, the measurement is redefined as

$$\begin{aligned} \mathbf{y}_{\text{trans}} &= \tilde{\mathbf{C}}^{i\,T}_{b'}\boldsymbol{\delta}\mathbf{v}^i = \tilde{\mathbf{C}}^{i\,T}_{b'}\left(\tilde{\mathbf{v}}^i - \mathbf{v}^i\right) = -\mathbf{d}\mathbf{v}^b \\ &= \underbrace{\begin{bmatrix}\mathbf{0}_{3\times3} & -\mathbf{I}_{3\times3} & \mathbf{0}_{3\times9}\end{bmatrix}}_{\mathbf{H}_{\mathbf{Lse},\text{trans}}}\mathbf{d}\mathbf{x}_{\mathbf{Lse}} \end{aligned} \tag{92}$$

It is shown that the redefined measurement transition matrix is now independent of the global state. In contrast, the measurement transition matrix $\mathbf{H}_{\mathbf{Rse}}$ is dependent on the global state $\tilde{\mathbf{v}}^i$ and cannot be transformed into state-independent form through linear operation. Such dependence will degrade the initial alignment performance with large misalignment.

For error state models based on $\mathbb{SO}(3)$, the process models (39) and (47) are both functions of the SINS calculated attitude and therefore, their performance will also be degraded due to nonlinearity caused by large misalignments. Since the velocity is a left invariant observation, initial alignment with model (47) will outperform that with (39), although they have the same measurement model.

***Remark 3:*** In the above discussion, we have presented the theoretical explanations for the fact that why the linear error state model can be used to accomplish the nonlinear initial alignment (even with extreme large misalignment). Next we will present another demonstration how the nonlinear attitude error can be exactly recovered from the linear attitude error model. Much of demonstration follows the works [28, 45].

With the left error definition (or body frame error for attitude), the corresponding attitude error equation is given by

$$\dot{\boldsymbol{\varphi}} = -\boldsymbol{\omega}^b_{ib} \times \boldsymbol{\varphi} \tag{93}$$

In (93) we have ignored the gyroscope measurement error $\boldsymbol{\delta\omega}^b_{ib}$, which will be facilitated to the following demonstration. It is known that when deriving (93), the first-order approximation of the attitude error matrix $\boldsymbol{\delta}\mathbf{C} = \mathbf{I}_{3\times3} + (\boldsymbol{\varphi}\times)$ has been used. The left error corresponding to $\mathbf{C}^i_b$ is given by

$$\boldsymbol{\delta}\mathbf{C} = \tilde{\mathbf{C}}^{i\,T}_b\mathbf{C}^i_b \tag{94}$$

The corresponding differential equation of $\boldsymbol{\delta}\mathbf{C}$ is given by

$$\begin{aligned} \boldsymbol{\delta}\dot{\mathbf{C}} &= \tilde{\mathbf{C}}^{i\,T}_b\dot{\mathbf{C}}^i_b + \dot{\tilde{\mathbf{C}}}^{i\,T}_b\mathbf{C}^i_b \\ &= \tilde{\mathbf{C}}^{i\,T}_b\mathbf{C}^i_b\left(\boldsymbol{\omega}^b_{ib}\times\right) - \left(\boldsymbol{\omega}^b_{ib}\times\right)\tilde{\mathbf{C}}^{i\,T}_b\mathbf{C}^i_b \\ &= \boldsymbol{\delta}\mathbf{C}\left(\boldsymbol{\omega}^b_{ib}\times\right) - \left(\boldsymbol{\omega}^b_{ib}\times\right)\boldsymbol{\delta}\mathbf{C} \end{aligned} \tag{95}$$

Let $\boldsymbol{\delta}\mathbf{C}_0 = \exp(\boldsymbol{\varphi}_0)$ be the initial left invariant error. It can be checked that

$$\boldsymbol{\delta}\mathbf{C}_t = \mathbf{C}^{i\,T}_{b,t}\boldsymbol{\delta}\mathbf{C}_0\mathbf{C}^i_{b,t} \tag{96}$$

is a solution to the error dynamics equation (95) through

$$\begin{aligned} \boldsymbol{\delta}\dot{\mathbf{C}}_t &= \mathbf{C}^{i\,T}_{b,t}\boldsymbol{\delta}\mathbf{C}_0\mathbf{C}^i_{b,t}\left(\boldsymbol{\omega}^b_{ib,t}\times\right) - \left(\boldsymbol{\omega}^b_{ib,t}\times\right)\mathbf{C}^{i\,T}_{b,t}\boldsymbol{\delta}\mathbf{C}_0\mathbf{C}^i_{b,t} \\ &= \boldsymbol{\delta}\mathbf{C}_t\left(\boldsymbol{\omega}^b_{ib,t}\times\right) - \left(\boldsymbol{\omega}^b_{ib,t}\times\right)\boldsymbol{\delta}\mathbf{C}_t \end{aligned} \tag{97}$$

Denote the vector attitude error corresponding to $\boldsymbol{\delta}\mathbf{C}_t$ as $\boldsymbol{\varphi}_t$. According to the Lie algebra for $\mathbb{SO}(3)$, (96) can be reorganized as

$$\exp(\boldsymbol{\varphi}_t) = \mathbf{C}^{i\,T}_{b,t}\exp(\boldsymbol{\varphi}_0)\mathbf{C}^i_{b,t} = \exp\left(\mathbf{C}^{i\,T}_{b,t}\boldsymbol{\varphi}_0\right) \tag{98}$$

It can be extracted from (98) that

$$\boldsymbol{\varphi}_t = \mathbf{C}^{i\,T}_{b,t}\boldsymbol{\varphi}_0 \tag{99}$$

The differential equation of (98) is given by

$$\dot{\boldsymbol{\varphi}}_t = -\left(\boldsymbol{\omega}^b_{ib,t}\times\right)\mathbf{C}^{i\,T}_{b,t}\boldsymbol{\varphi}_0 = -\boldsymbol{\omega}^b_{ib,t}\times\boldsymbol{\varphi}_t \tag{100}$$

which is just the equation (93). It is shown that from (94) to (100), we have not made the first-order approximation for the attitude error. That is to say if the initial error is known, the nonlinear error dynamics can be exactly recovered from this linear system. This property is known as Log-Linear property. In other words, when deriving the linear attitude error model, the small attitude error assumption surprising does not lost accuracy.

It is shown that the aforementioned feature is also suitable for the traditional $\mathbb{SO}(3)$ based model (44a) with body frame attitude error. Unfortunately, in the traditional model (44), only the attitude error is expressed in body frame, the velocity and position errors are still expressed in inertial frame. It can be said that the expressed frame is not consistent for model (44). This is the main reason for the performance degradation of this model.

## VIII. Simulation Results

This section is devoted to numerically evaluating the initial alignment performance based on different error state models. Specially, the following four algorithms are evaluated for comparison:

Initial alignment based on error state-space model (38) and (50), denoted as RSO-KF.

Initial alignment based on error state-space model (46) and (50), denoted as LSO-KF.

Initial alignment based on error state-space model (63) and (67), denoted as RSE-KF.

Initial alignment based on error state-space model (77) and

(81), denoted as LSE-KF.

The SINS under investigation is located at medium latitude $34^\circ$ that is equipped with a triad of gyroscopes (drift $0.01^\circ/h$, noise $0.001^\circ/\sqrt{h}$ ) and accelerometers (bias $100\mu g$ , noise $10\mu g/\sqrt{Hz}$ ). The SINS is assumed to be under static condition and its sampling rate is $100Hz$ . Firstly, the four algorithms are evaluated by implementing 200 Monte Carlo runs. The horizontal misalignments are assumed to satisfy the uniform distribution within $\left[-90^\circ\ 90^\circ\right]$ and the heading misalignment is assumed to satisfy the uniform distribution within $\left[-180^\circ\ 180^\circ\right]$. The initial filtering parameters are all assumed to be same for the four algorithms. Specially, the initial error states are all assumed to be zero. The initial covariance is diagonal with attitude errors elements set to $(\boldsymbol{\varphi}/3)^2$ with $\boldsymbol{\varphi}$ denoting the corresponding misalignments.

It should be noted that different attitude errors definitions are only used to define different state errors which will be estimated using KF. After the KF estimation, the SINS calculated attitude results will be refined using the estimated error state according to their definitions, respectively. That is to say, different attitude error definitions have different refinement forms. After the refinement, we can get different attitude estimates by different initial alignment methods. For comparison presentation, the attitude errors are all given by $\boldsymbol{\delta\alpha} = \boldsymbol{\alpha}_{\text{true}} - \hat{\boldsymbol{\alpha}}_{\text{estimate}}$ directly. This is the common procedures for comparison of different initial alignment methods.

The attitude estimate errors by the four initial alignment algorithms across 200 Monte Carlo runs are shown in Figs. 2-4, respectively. It is clearly shown that the initial alignment methods with left error models outperform those with right error models. This is because that the velocity is a left-invariant type observation. Moreover, when the initial misalignment is large, RSO-KF cannot converge.  Although RSE-KF can accomplish the initial alignment, its performance is much worse than LSO-KF and LSE-KF in terms of both convergent speed and steady-state accuracy. This is because that the measurement model for RSE-KF is dependent on the SINS calculated velocity. According to Lie group theory, the nonlinear group state cannot be recovered exactly from the corresponding linear error state vector. The LSO-KF can also accomplish the initial alignment even with very large misalignments, although its process model is dependent on the SINS calculated attitude. Its performance is even better than RSE-KF, which means that for this simulation scenario the state-dependence in measurement model (67) degrades the initial alignment performance more severely than that in process model (46). From Fig. 2-4, we can see that the LSO-KF and LSE-KF have a very similar convergent speed. The steady-state estimate results by the two methods are shown in Fig. 5. It is clearly shown that LSE-KF performs much better than LSO-KF. That is to say, the state-dependence in process model (46) still has a negative effect on the initial alignment performance with large misalignment. In contrast, the LSE-KF can accomplish the initial alignment quite well even with extreme large misalignments. This is consistent with the theoretical analysis in Section VII. Therefore, LSE-KF can be used directly in practical initial alignment applications even with no prior knowledge on the initial attitude. In other words, there is no need to perform coarse alignment for LSE-KF and therefore, the total alignment time is expected to be reduced.

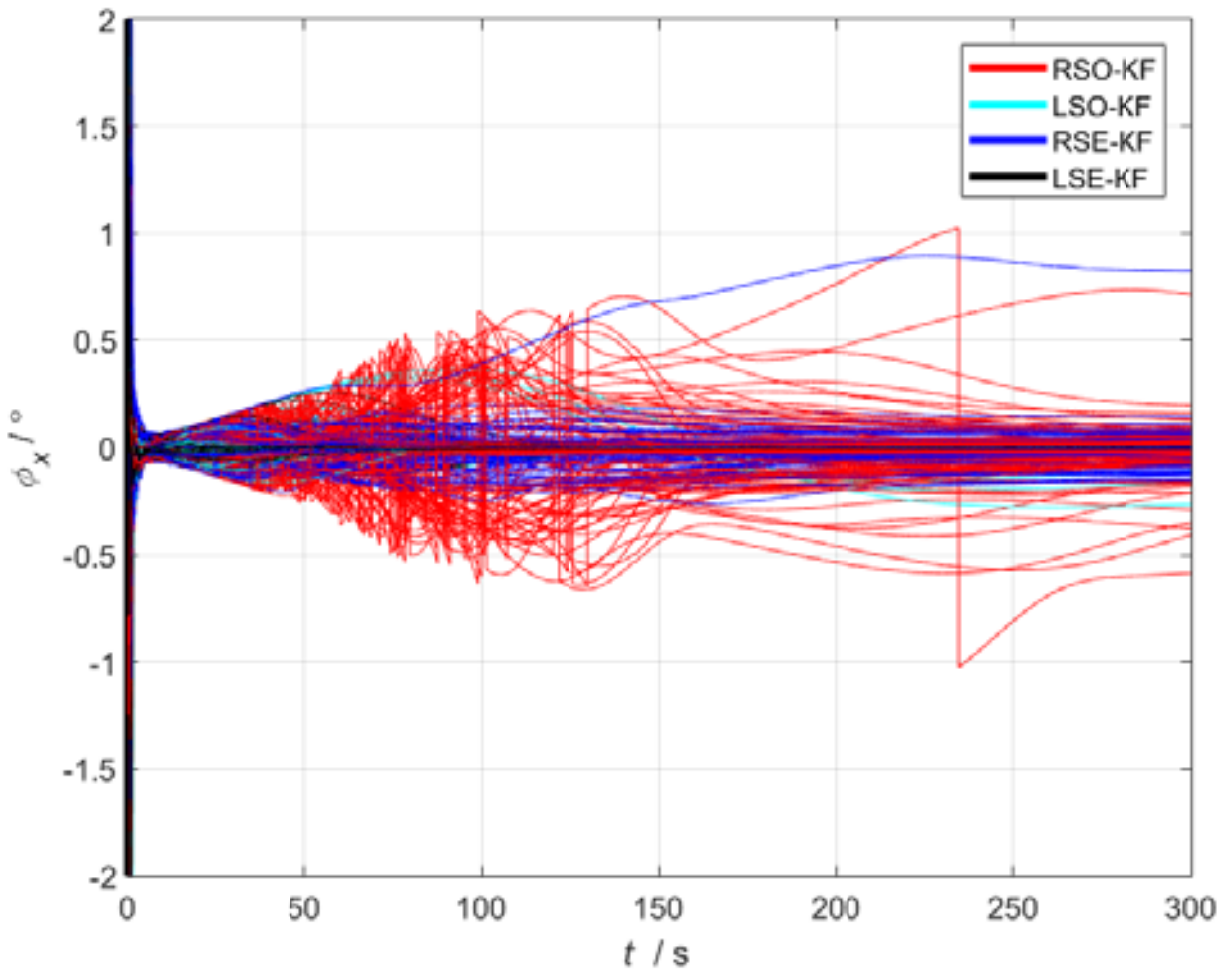

Fig. 2. Pitch angles estimate errors across 200 Monte Carlo runs

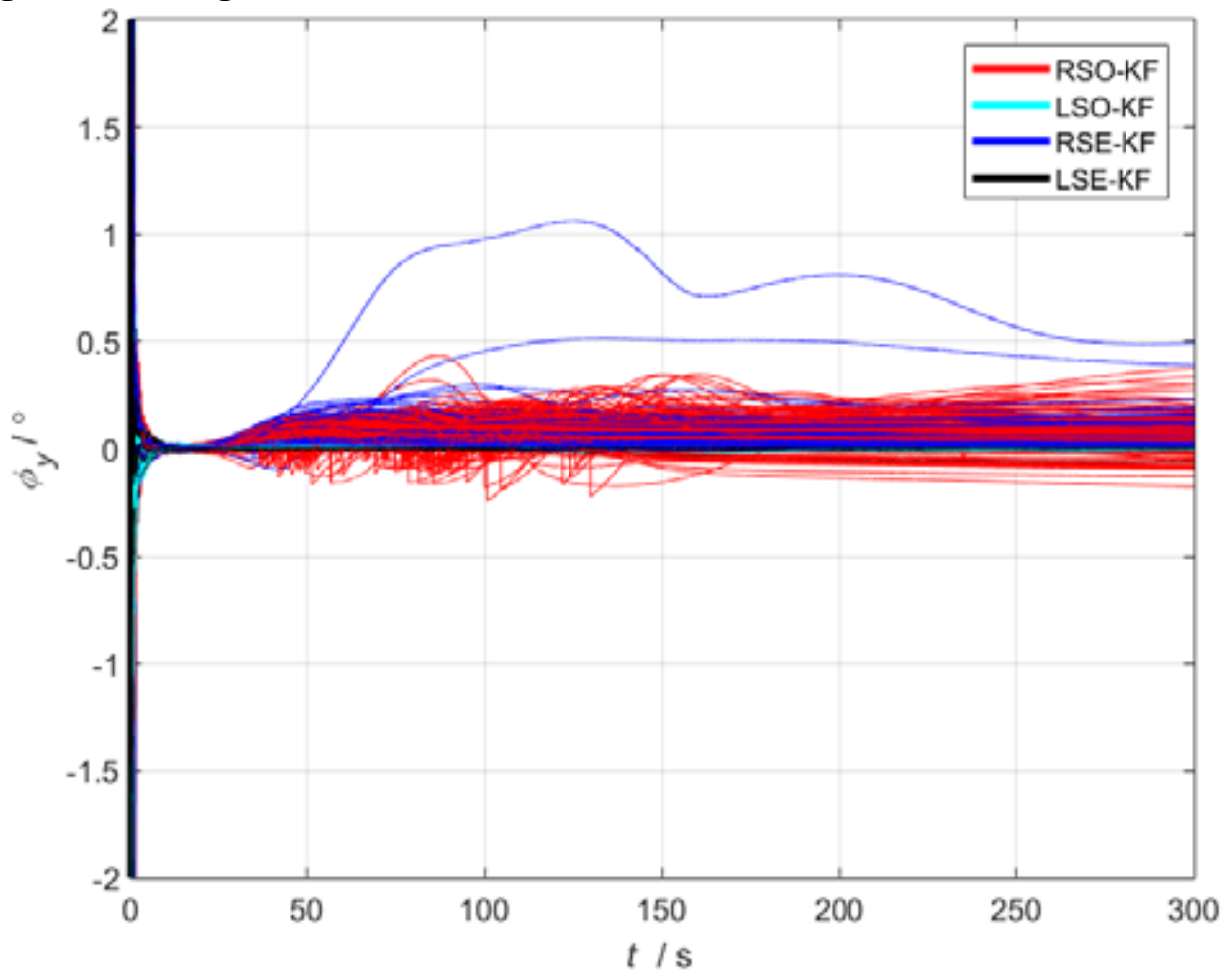

Fig. 3. Roll angles estimate errors across 200 Monte Carlo runs

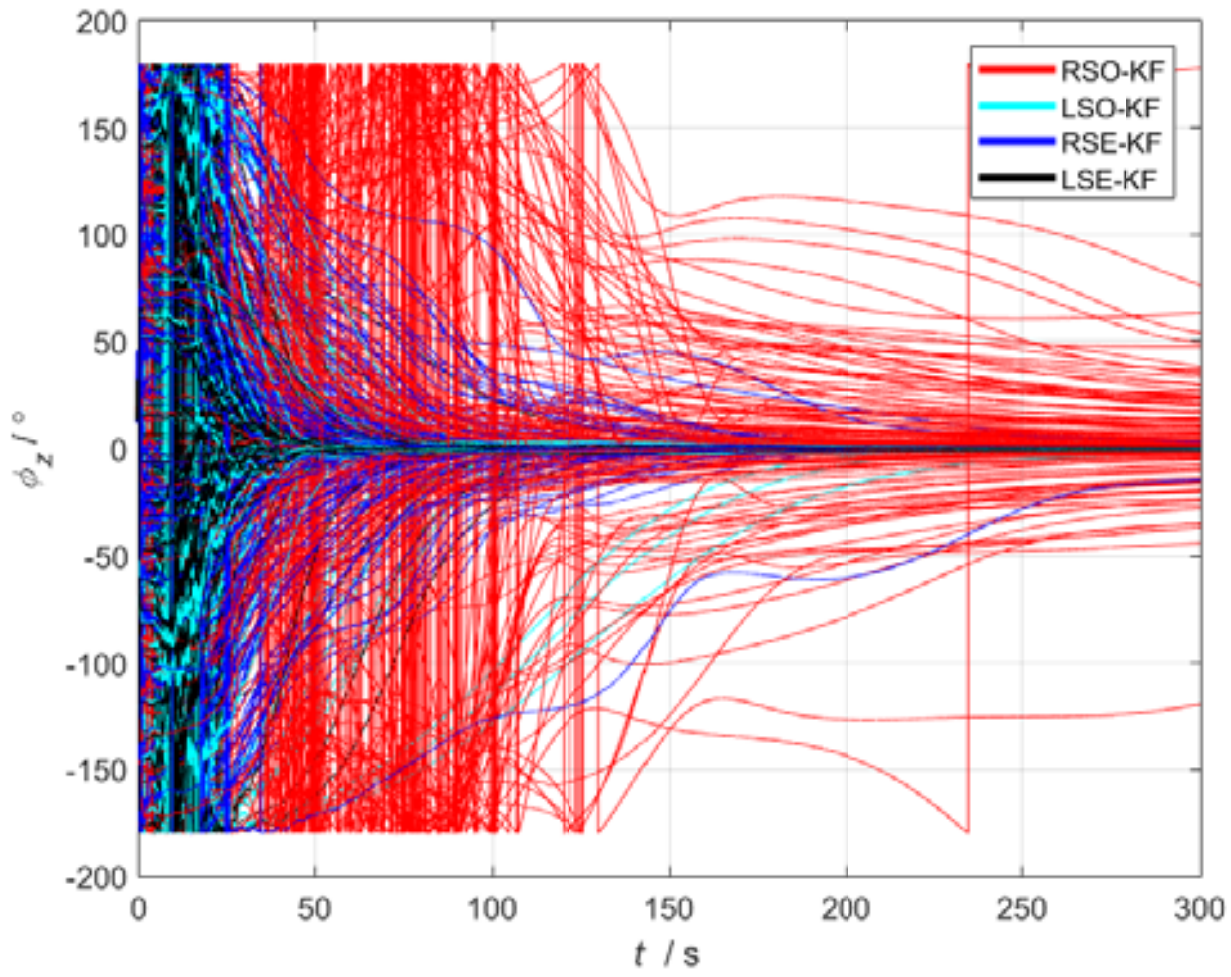

Fig. 4. Yaw angles estimate errors across 200 Monte Carlo runs

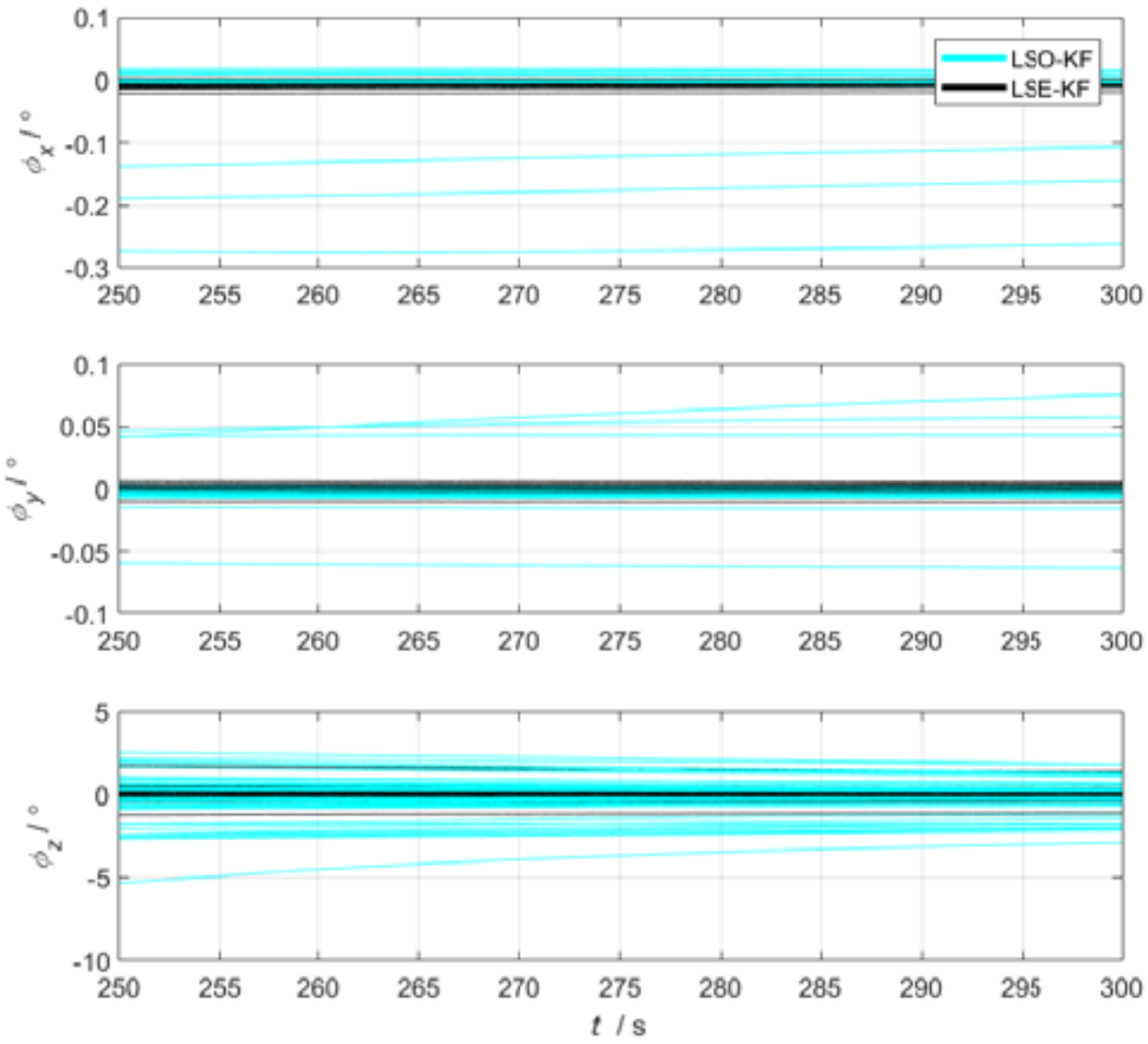

Fig. 5. Steady-state estimate errors by LSO-KF and KSE-KF

In order to further evaluate the performance of these four initial alignment algorithms, simulations with different yaw misalignment angles have been carried out. The resulting yaw angles estimate errors by the four algorithms are shown in Fig. 6 and 7. It is shown that the convergence of the RSO-KF deteriorates much faster with the increase of the initial yaw misalignment. These results highlight the necessary of coarse alignment stage for the traditional RSO-KF based fine alignment. In contrast, LSO-KF, RSE-KF and LSE-KF all can converge even with large misalignments. Moreover, LSO-KF and LSE-KF converge faster than RSE-KF. However, the steady-state estimates of LSE-KF are more accurate than those of LSO-KF. These results also indicate that the left error model based on $\mathbb{SE}_2(3)$ is more like a log-linear error system, through which the nonlinear estimation error (for the case with large misalignments) can be recovered more accurately than that by other models.

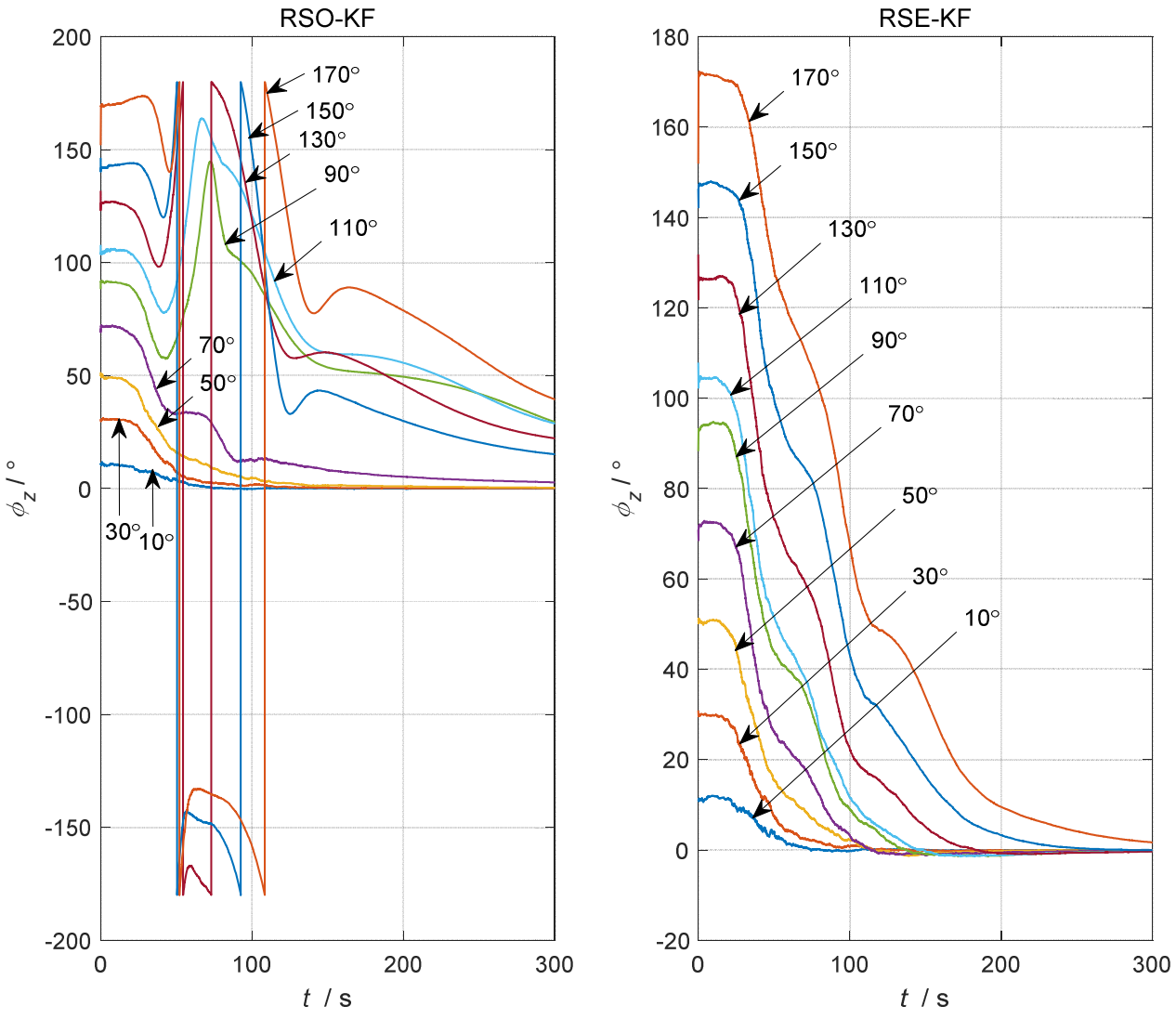

Fig. 6. Yaw angles estimate errors by RSO-KF and RSE-KF

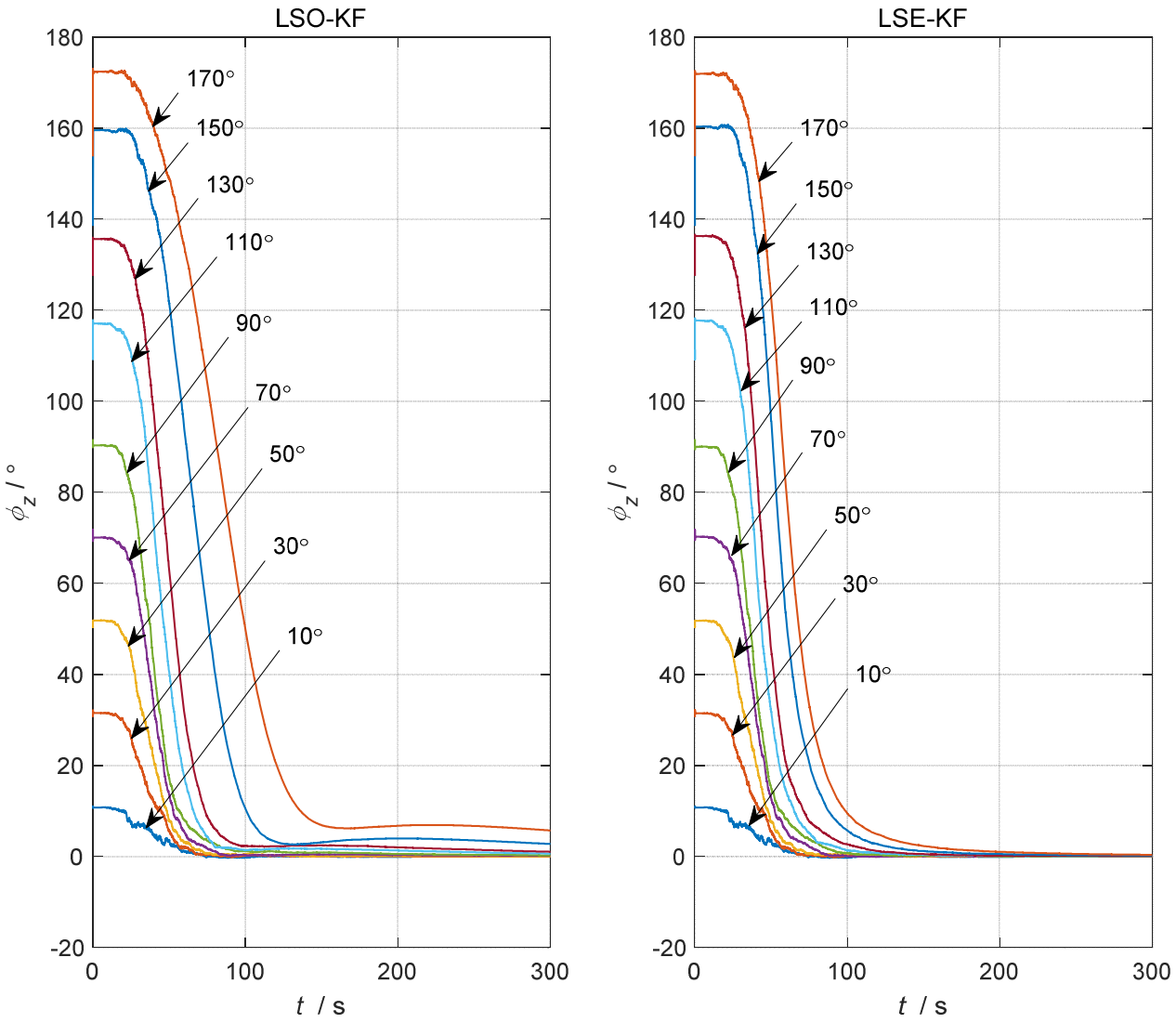

Fig. 7. Yaw angles estimate errors by LSO-KF and LSE-KF

For initial alignment with large misalignments, some nonlinear initial alignment methods have been investigated. Therefore, one would like to compare the performance of LSE-KF with nonlinear initial alignment methods. With this consideration, the unscented Kalman filtering (UKF) with nonlinear Euler angle error model in [16] which is argued to be able to handle large initial misalignments is evaluated as comparison. The simulation setting is the same with the above simulation evaluation. The initial alignment results by LSE-KF and UKF across 200 Monte Carlo runs are shown in Fig. 8-10, respectively. It is clearly shown that the LSE-KF has a much faster convergent speed and the steady state is also more accurate and consistent. Moreover, LSE-KF is based on linear KF and therefore, its computational burden is much less than that of UKF. This MC evaluation demonstrates that LSE-KF can accomplish the initial alignment with extreme large misalignments and its performance is even better than the nonlinear initial alignment method in terms of both convergent speed and steady state accuracy.

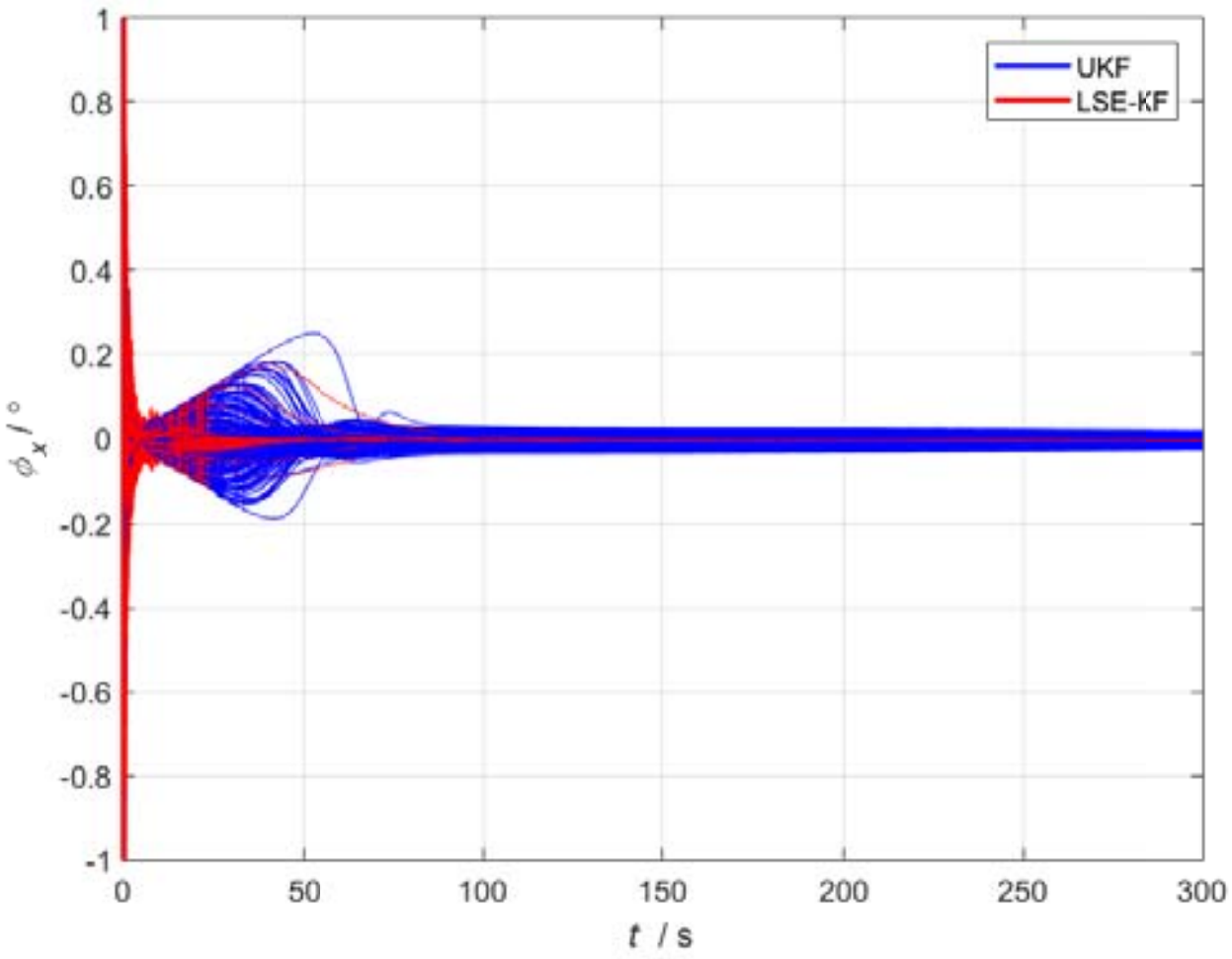

Fig. 8. Pitch angles estimate errors by LSE-KF and UKF

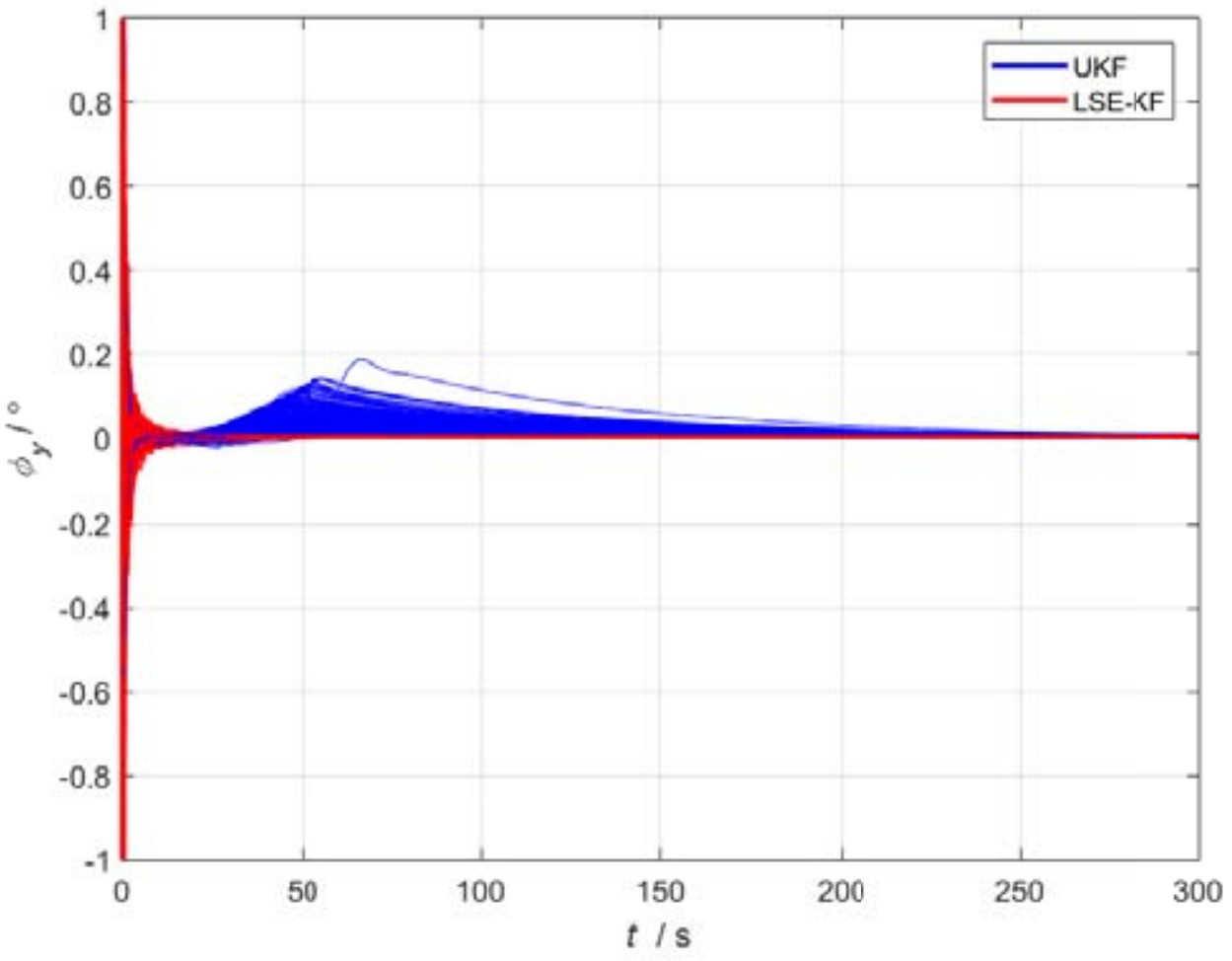

Fig. 9. Roll angles estimate errors by LSE-KF and UKF

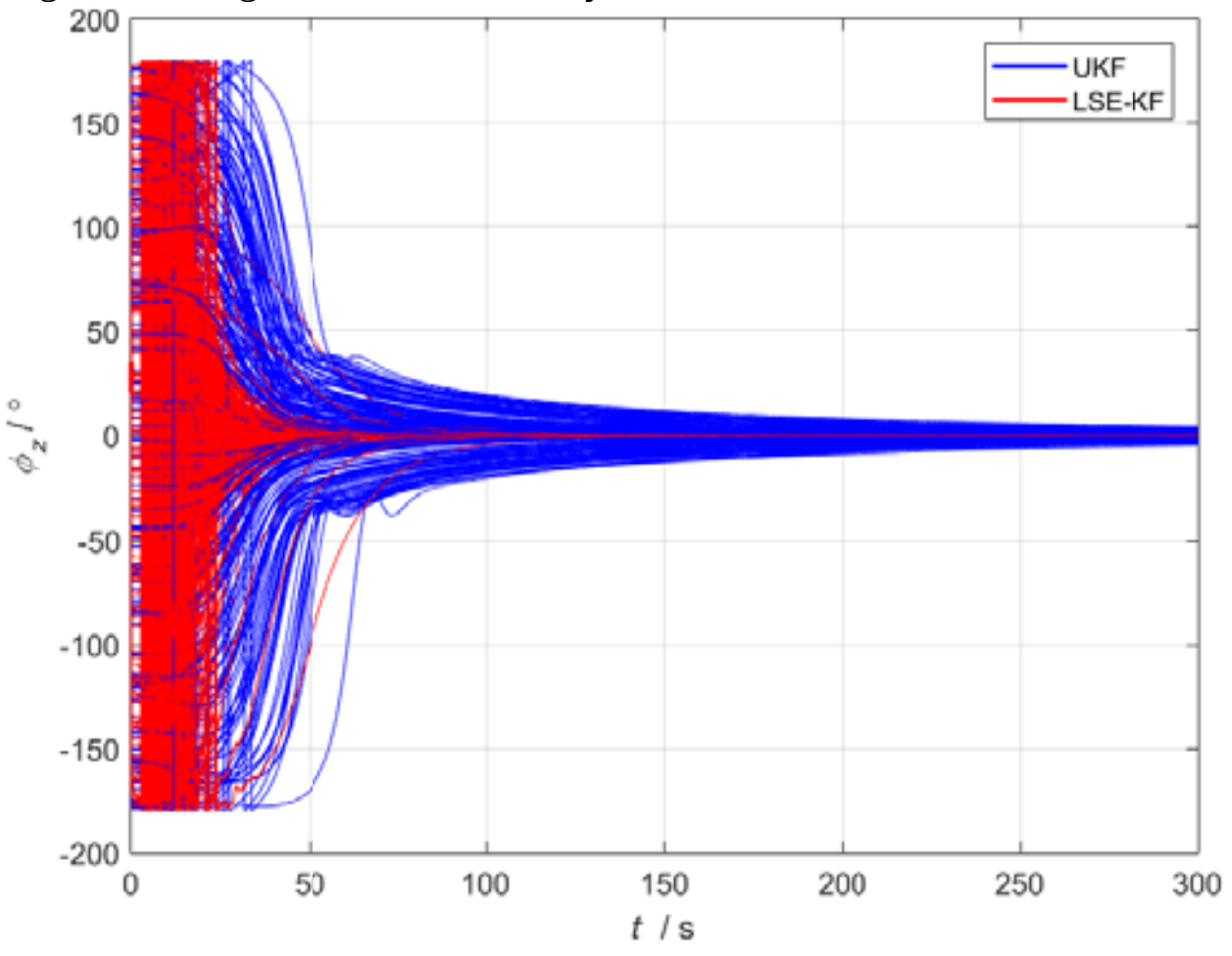

Fig. 10. Yaw angles estimate errors by LSE-KF and UKF

Besides the MC evaluation, a single initial alignment run is carried out with misalignment $\begin{bmatrix}89 & 89 & 179\end{bmatrix}^{\circ}$. In this evaluation, the utmost precision of the static initial alignment is also calculated as comparison reference. For the given specifications of the inertial sensors, the utmost precision of the static initial alignment is given by

$$\begin{aligned} \boldsymbol{\varphi}_x &= -\nabla_N / g \\ \boldsymbol{\varphi}_y &= \nabla_E / g \\ \boldsymbol{\varphi}_z &= \tan L \cdot \nabla_E / g - \boldsymbol{\varepsilon}_E / \left(\omega_{ie} \cos L\right) \end{aligned} \tag{101}$$

where $\nabla_E$ and $\nabla_N$ are the east and north components of the accelerometers drift bias and $\boldsymbol{\varepsilon}_E$ is the east component of the gyroscopes drift bias. The alignment results by the two methods coupled with the utmost precision are shown in Fig. 11-13, respectively. The superiority of LSE-KF over UKF can also be clearly observed. Meanwhile, the steady-state estimate of LSE-KF is also very close to the utmost precision. That is to say, even with very large misalignment, LSE-KF can still accomplish the initial alignment effectively.

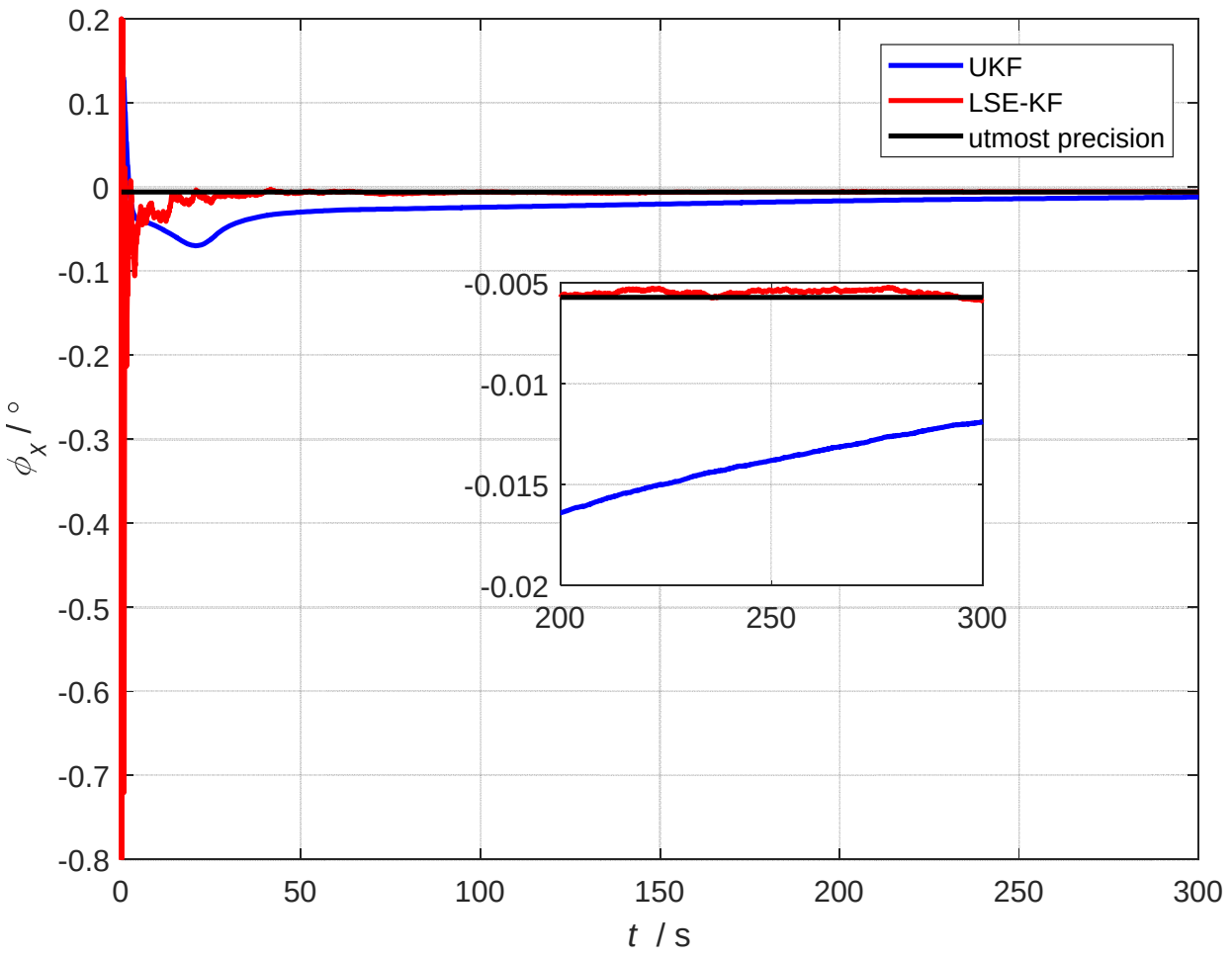

Fig. 11. Pitch angle alignment errors by LSE-KF and UKF with extreme large misalignments

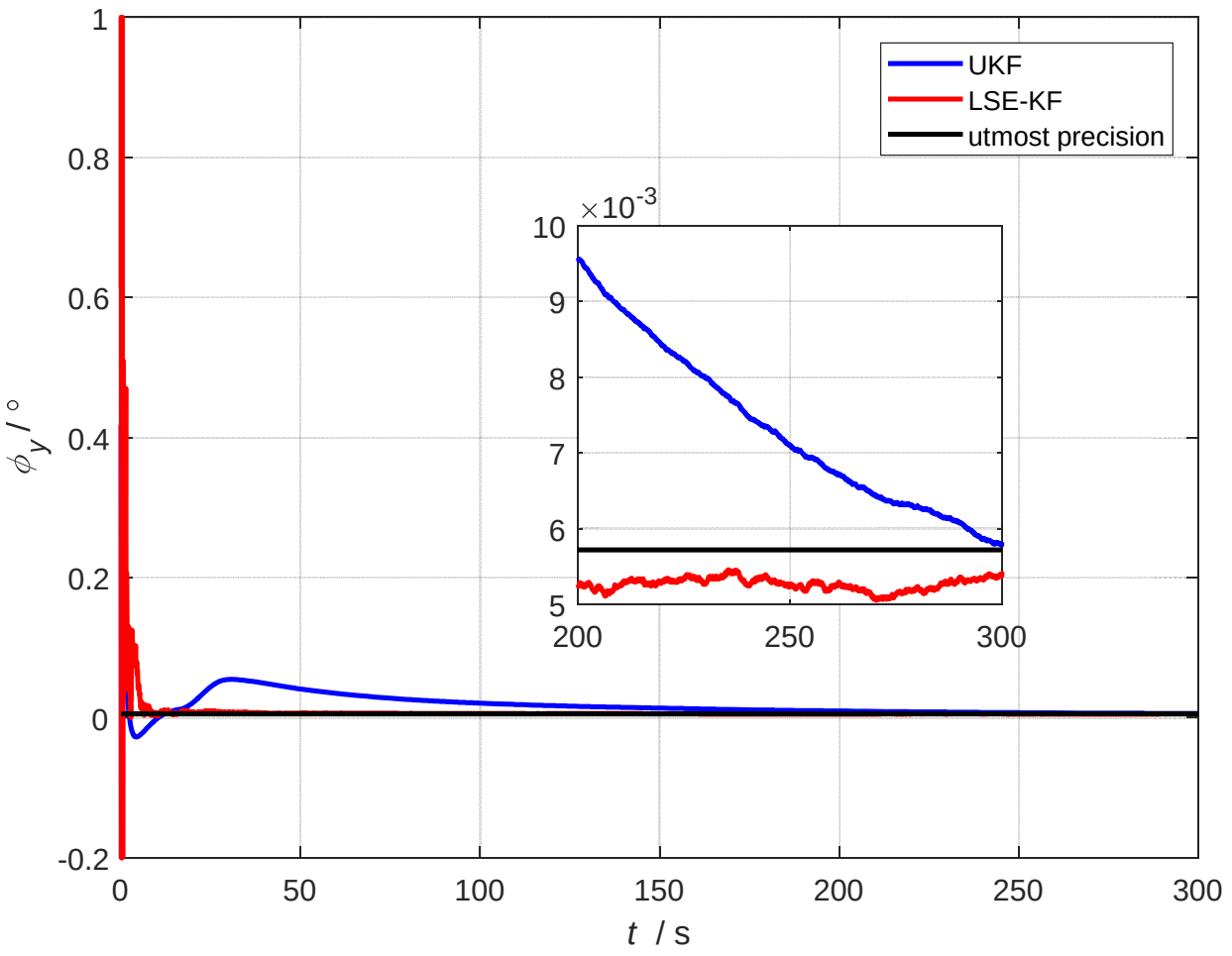

Fig. 12. Roll angle alignment errors by LSE-KF and UKF with extreme large misalignments

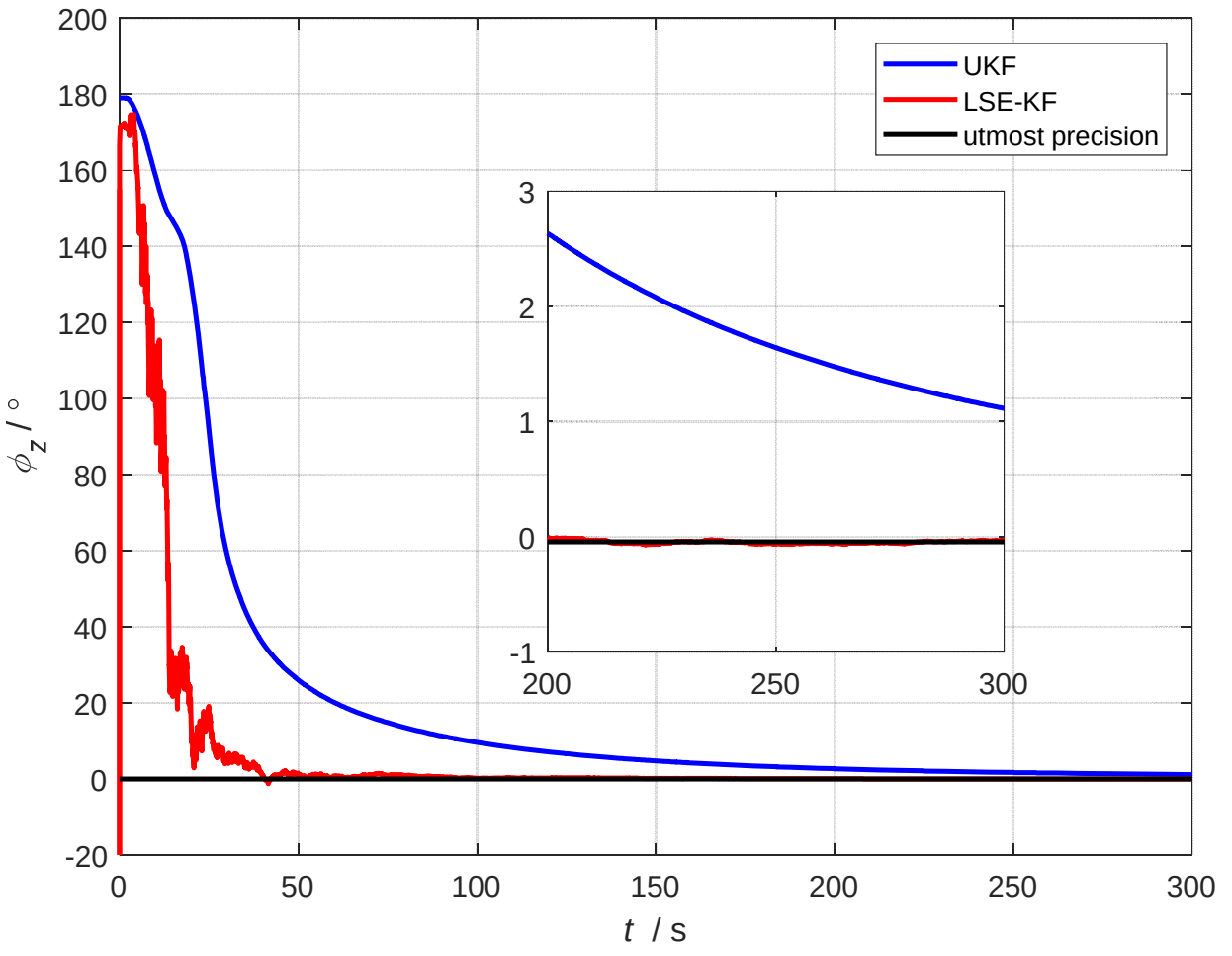

Fig. 13. Yaw angle alignment errors by LSE-KF and UKF with extreme large misalignments

## IX. Field Test Results

In this section, the aforementioned four linear initial alignment algorithms are further evaluated using two field test data segments. One was collected from a navigation-grade ring laser SINS and the other was collected from intermediate-grade fiber optic SINS.

## A. *Ring Laser SINS Field Test*

A navigation-grade SINS equipped with a triad of ring laser gyroscopes and accelerometers is fixed inside the car. The SINS sampling rate is 125*Hz*. The dominating inertial sensor errors for the SINS include: the gyroscope constant drift: $0.007^\circ/h$, the accelerator bias: $50\mu g$ .

Since for the navigation-grade SINS, there is no attitude reference for comparison, we have designed the following experiment sequences: after the SINS is powered-on, some peoples get on and off the car frequently to enhance the disturbance intentionally and a test segment of about 300-seconds test data is collected to test the four linear initial alignment algorithms. The OBA in [8] is carried out using this data segment to obtain the attitude at the very start. The attitude obtained through the OBA is used as the initial value for the RSO-KF, denoted as "RSO-KF with small misalignment" hereafter. It should be noted that the performance of the traditional RSO-KF with small initial misalignment has been widely approved in commercial products. For a harsh comparison of four evaluated alignment algorithms, the initial misalignment $\begin{bmatrix}5^\circ & 5^\circ & 100^\circ\end{bmatrix}$ is intentionally added to the attitude estimate by OBA.

The alignment results provided by the four methods using collected data under swaying condition are shown in Fig. 14-16, respectively. Parts of the three attitude results from 100s to 300s have been plotted in the right subfigures. For clarity, the yaw angle alignment result by RSO-KF has been omitted in the right subfigure of Fig. 16, because its result is far away from other results. It is shown that the LSE-KF can converge to the same value as the "RSO-KF with small misalignment" within a very short time period, although starting very far from the true value. In contrast, the RSE-KF needs more time to converge and its stead-state estimate is also apart from the results provided by "RSO-KF with small misalignment", especially for pitch and yaw angles estimate. That is to say, we cannot recover the nonlinear estimation error from the linear model (63). This is because that the right error model (63) and (67) is dependent on the global state and therefore does not possess the Log-Linear property. So, in practical applications, coarse alignment is still necessary for RSE-KF in order to guarantee accurate steady-state estimate. The LSO-KF has a similar convergent speed with LSE-KF. The steady-state accuracy of LSO-KF is also nearly the same with LSE-KF for pith and roll angles. However, for yaw angle estimate, the steady-state accuracy of LSO-KF is worse than that of LSE-KF. In this respect, the LSE-KF is still the best method within the four evaluated methods, which is consistent with the theoretical analysis and the simulation test results.

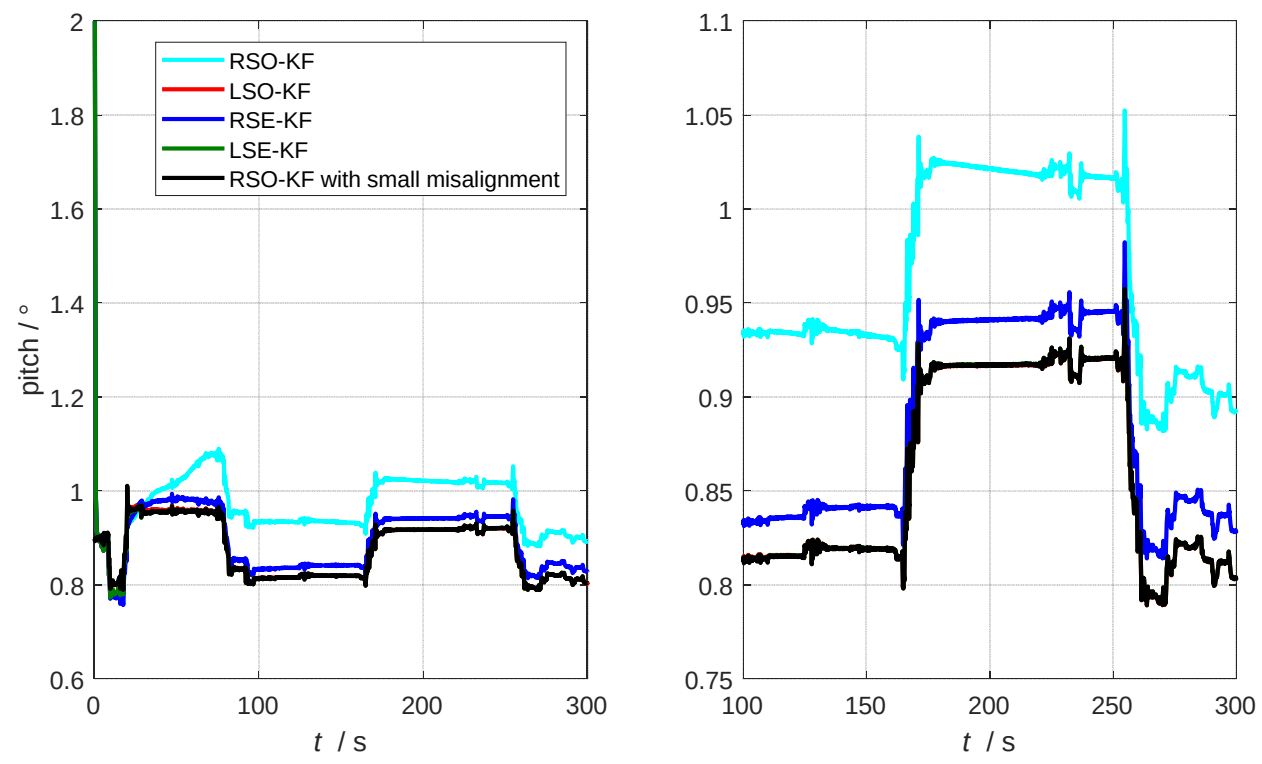

Fig. 14. Pitch alignment results for ring laser SINS

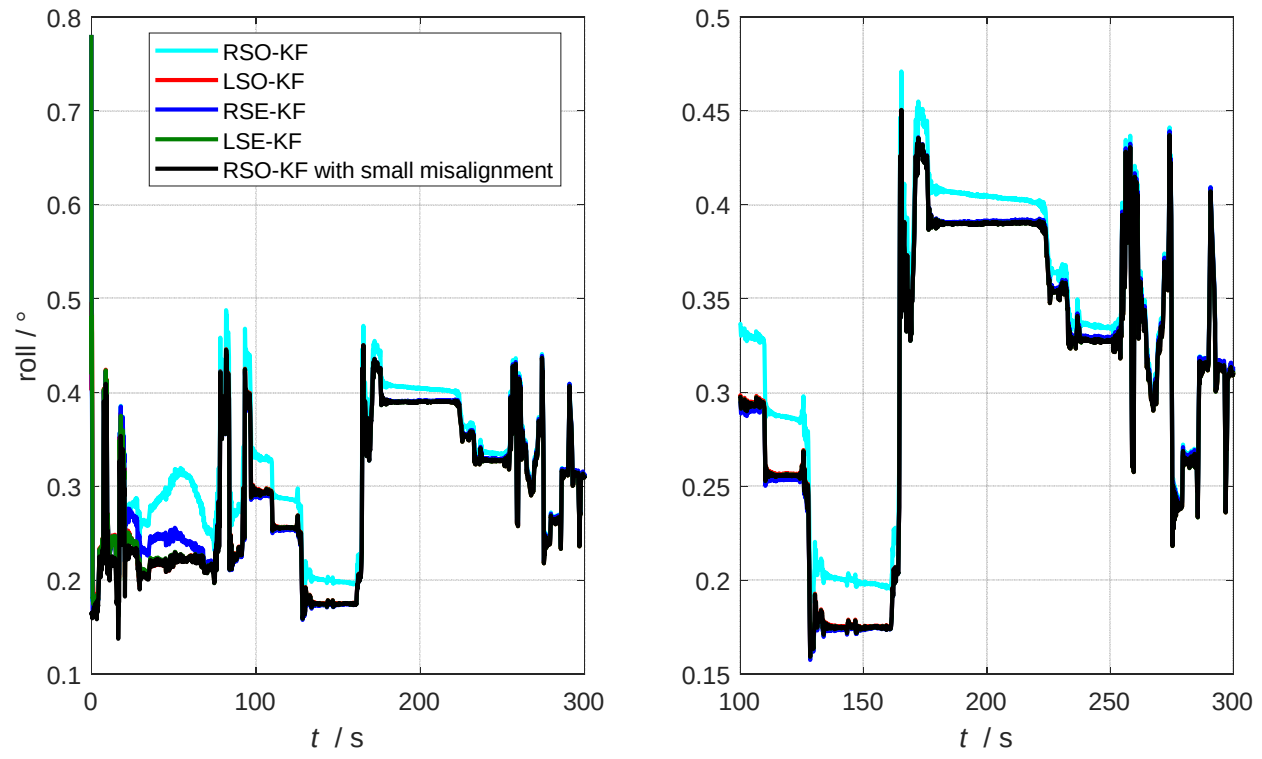

Fig. 15. Roll alignment results for ring laser SINS

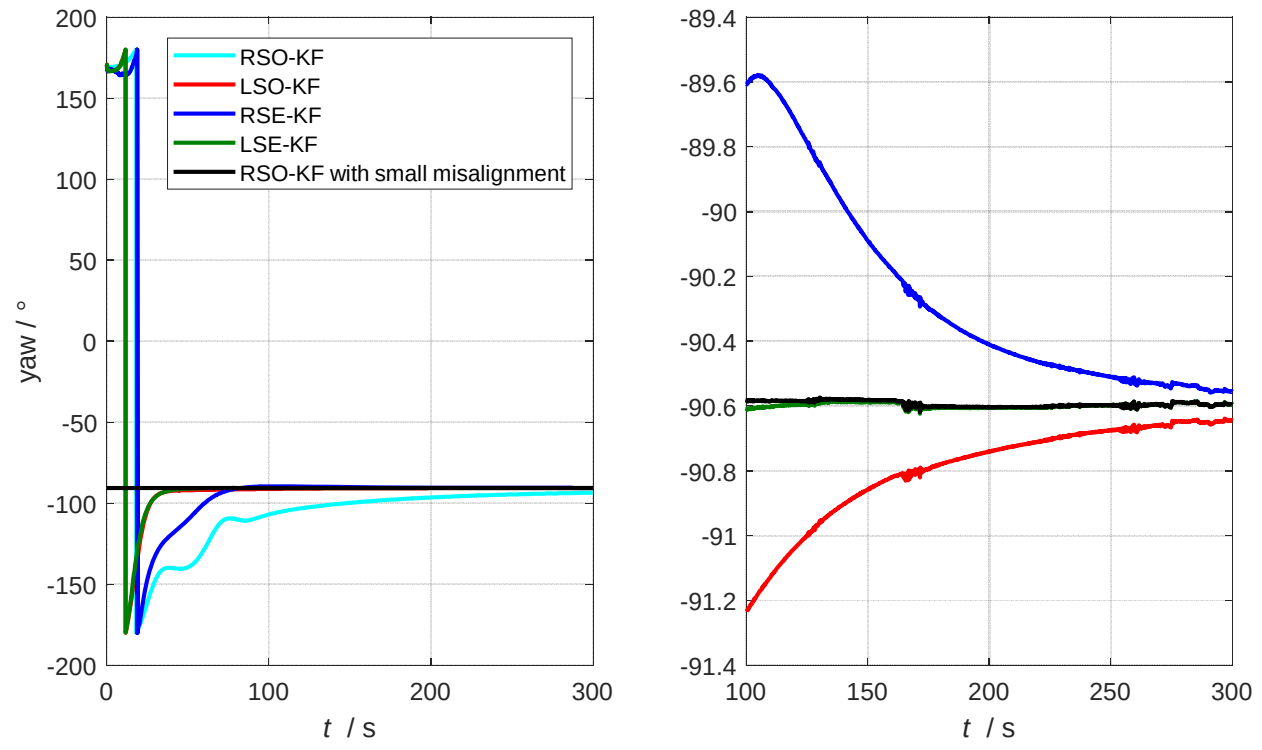

Fig. 16. Yaw alignment results for ring laser SINS

## B. *Fiber Optic SINS Field Test*

In this section, the four linear alignment algorithms are evaluated using dynamic data collected from an intermediate-grade fiber optic SINS. In this experiment, a POS, consisting of a navigation-grade fiber optic SINS and GNSS, can be used to provide attitude reference. The specifications of the navigation-grade fiber optic SINS are given as follows: drift bias of gyroscope is about $0.01^\circ/h$ and drift bias of accelerometer is about $20\mu g$ . The specifications of the intermediate-grade fiber optic SINS are given as follows: drift bias of gyroscope is about $0.3^\circ/h$ and drift bias of accelerometer is about $20mg$ . The SINS sampling rate is 200*Hz*. A data segment of 300 seconds was collected to evaluate the four linear alignment algorithms. The reference attitude is shown in Fig. 17. The reference velocity provided by the GNSS is shown in Fig. 18.

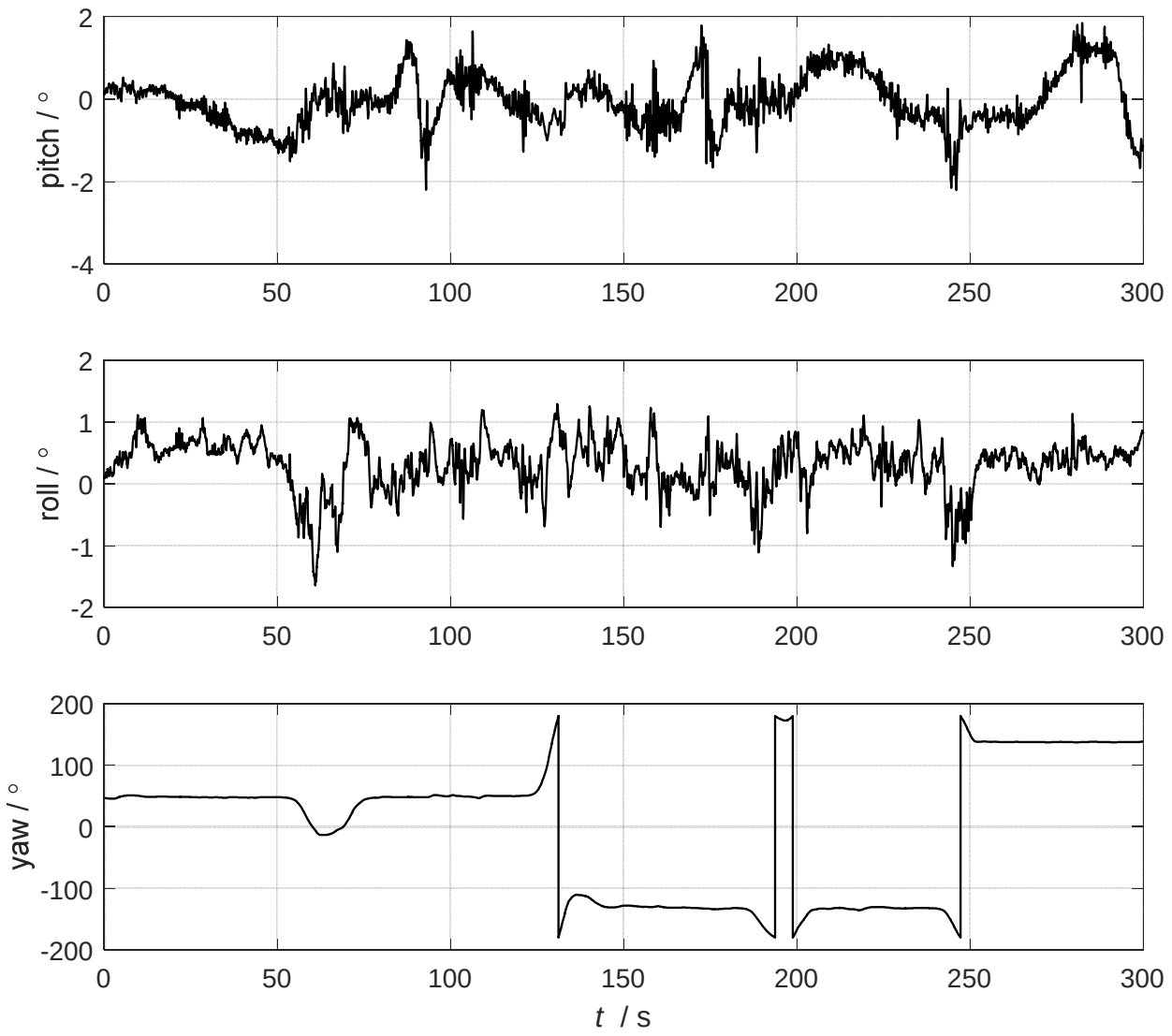

Fig. 17. Reference attitude during the experiment

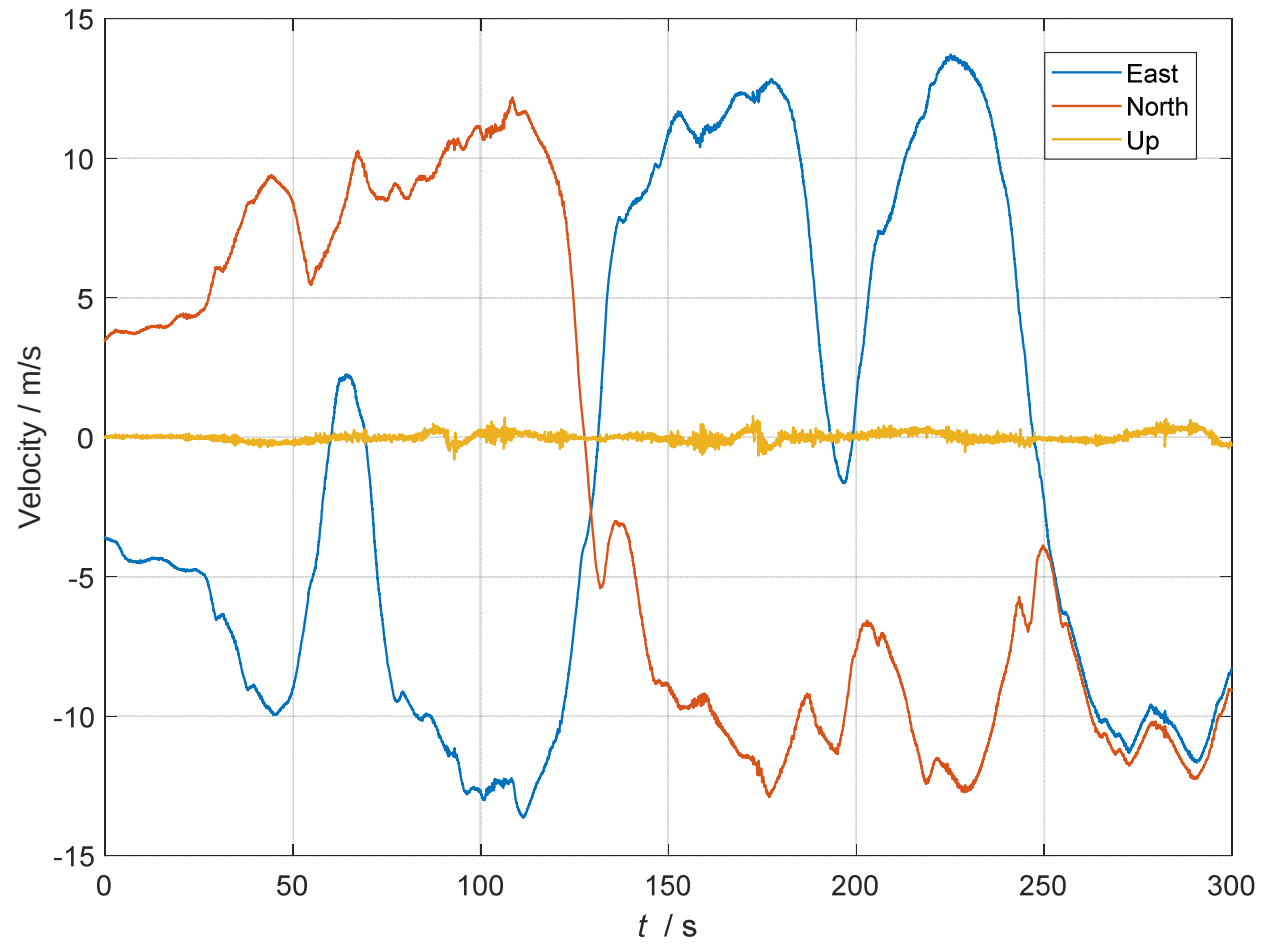

Fig. 18. Reference velocity during the experiment

In this field test, initial misalignment $\left[89^{\circ};89^{\circ};179^{\circ}\right]$ is added for the four evaluated methods and the corresponding alignment results are presented in Fig. 19-21, respectively. It is shown that, under dynamic conditions, all the four methods have a very fast convergent speed. The methods with right error models can also accomplish the initial alignment with extreme large misalignments. This is because that under dynamic conditions, the observability the three attitude angles can be strengthened [46]. The initial alignment methods with left error state models converge faster than those with right error state models. This is because that the measurement is left-invariant. In this field test, LSO-KF and LSE-KF perform nearly identical with each other in terms of both convergent speed and steady-state accuracy.

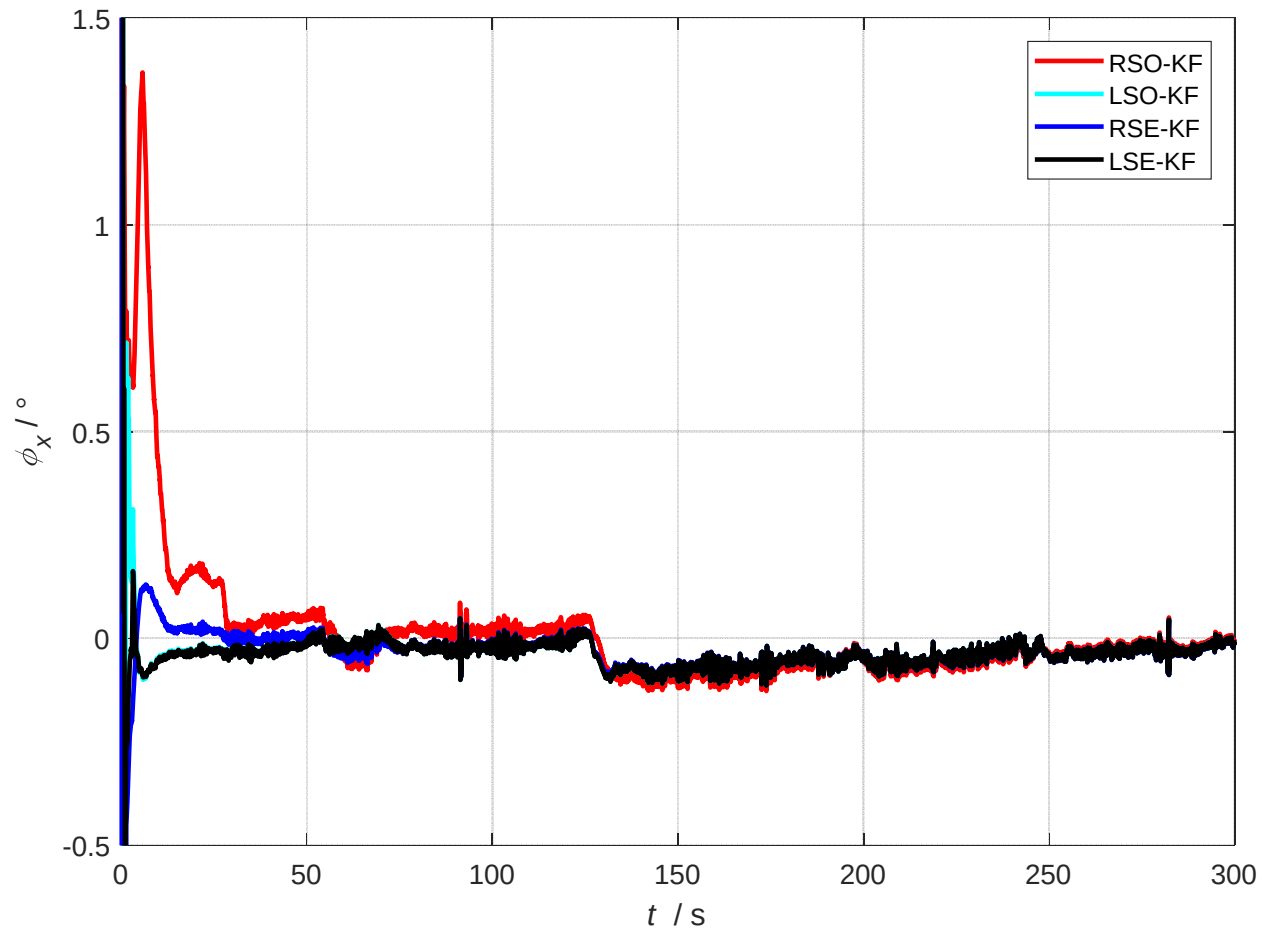

Fig. 19. Pitch alignment results for fiber optic SINS

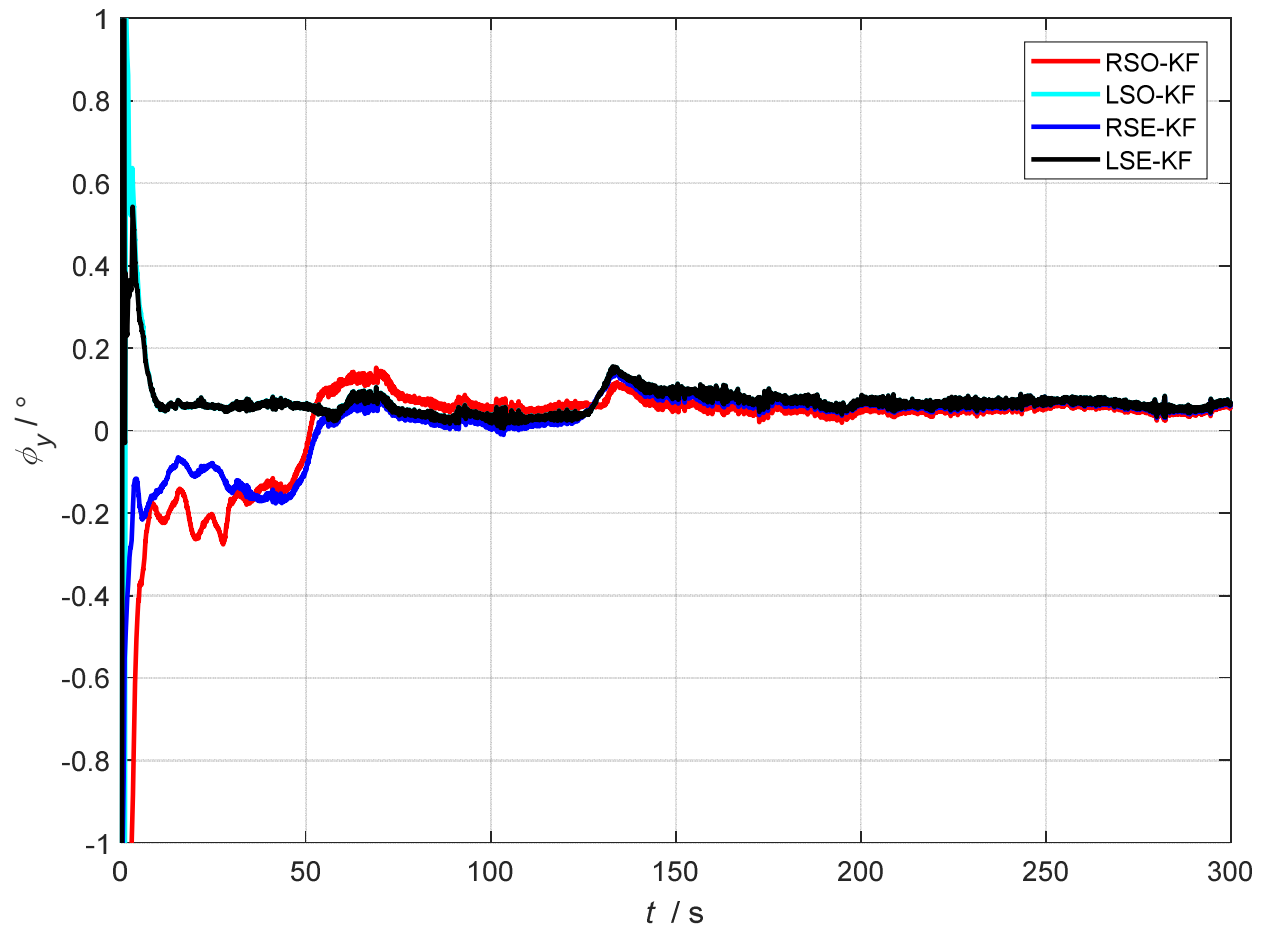

Fig. 20. Roll alignment results for fiber optic SINS

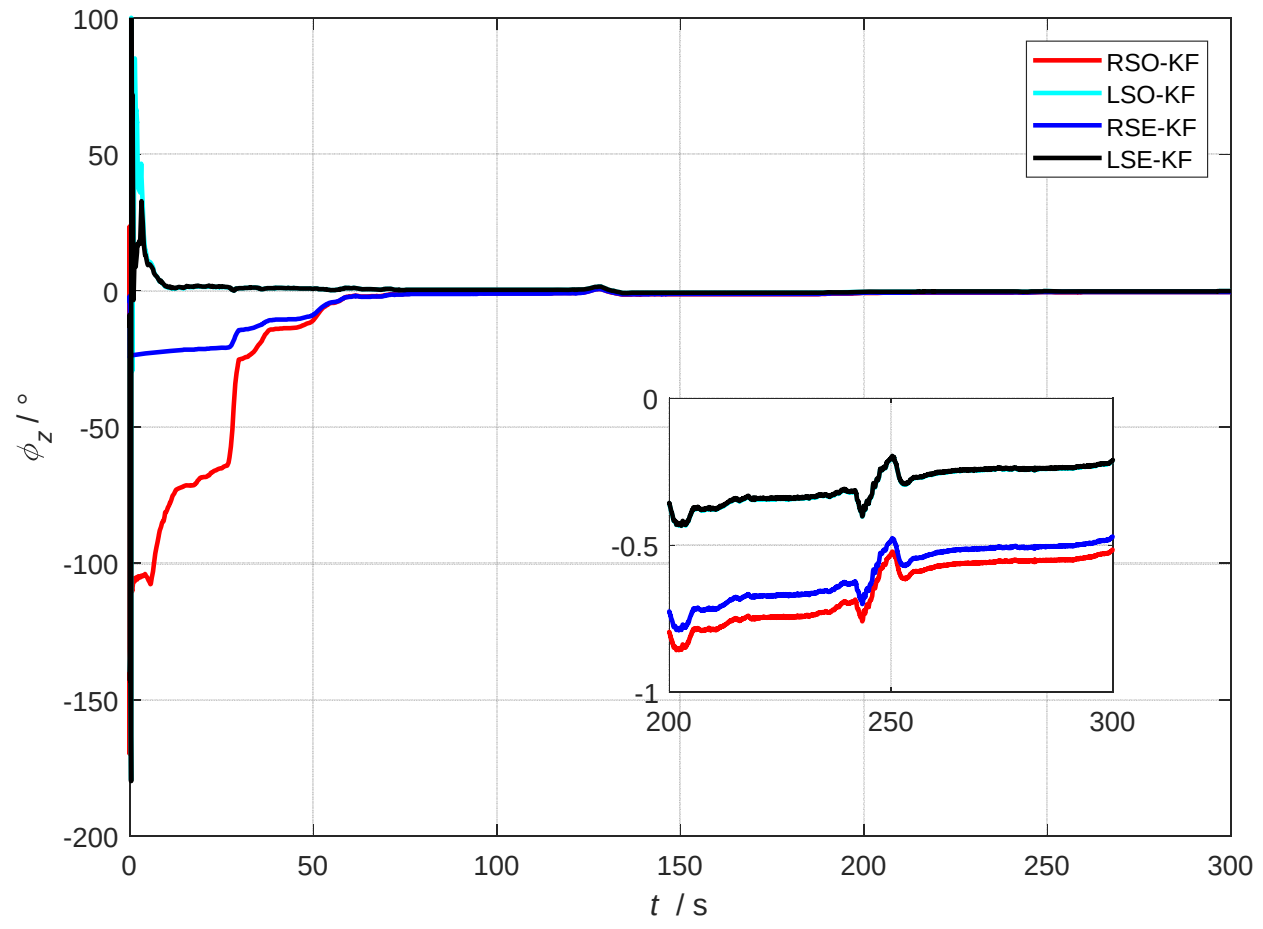

Fig. 21. Yaw alignment results for fiber optic SINS

## X. Conclusions

This paper investigates the $\mathbb{SE}_2(3)$ based KF for SINS initial alignment. Through introducing an artificially defined inertial frame, the desired attitude is decomposed into two parts, one of which can be calculated directly with accurate known initial values. The other part coupled with the velocity and position in the defined inertial frame are formulated as the SINS mechanization in inertia frame. Based on the reconstructed

SINS differential equations in inertial frame, both the right and left error state models are derived making use of the matrix Lie group theory. Compared with the traditional error state models, the $\mathbb{SE}_2(3)$ based models treat the attitude, velocity and position together as element of $\mathbb{SE}_2(3)$. Through such formulation, the defined group state model satisfies a group affine property and the corresponding linear error model satisfies a log-linear autonomous differential equation on the Lie algebra. It is proven that the nonlinear group error can be exactly recovered by such linear vector error model, which enables the linear KF based initial alignment with extreme large misalignments be possible. The explicit initial alignment procedures based on the derived error state models are also presented. Through extensive simulation and field test evaluations under different conditions, the results show that the LSE-KF can always performs quite well in terms of both convergent speed and steady-state estimate. Due to its ability to handle the large misalignments, the LSE-KF can be used directly without the traditional coarse alignment stage. In this respect, there will be no transitions between different alignment stages and can further improve the initial alignment speed. Moreover, the LSE-KF is still within the linear Kalman filtering framework and therefore it is very computational efficient. Therefore, the LSE-KF is promising in practical products.

In this paper, the initial alignment is accomplished aided by GNSS. In such scenario, the geographical velocity and position are known, which is just the prerequisite for the model decomposition. However, for SINS initial alignment aided by Doppler velocity log or Odometer [47-49], the proposed method is no longer suitable, because these aided sources can only provide velocity expressed in body frame. In the future work, SINS linear initial alignment aided by Doppler velocity log or Odometer with large misalignments will be investigated by means of Lie group theory.

## Acknowledgment

The authors would like to thank the authors of [44] for providing the open source codes in https://github.com/CAOR-MINES-ParisTech/ukfm.These programs have given the authors a more thoroughly understanding of the matrix Lie group based algorithms and accelerated the corresponding investigation in this paper. The first author would also like to thank Dr. Martin Brossard from MINES ParisTech, PSL Research University, for his kind guidance and help in investigating of the $\mathbb{SE}_2(3)$ based algorithms. The first author would also like to thank Yuanxin Wu and Wei Ouyang from Shanghai Key Laboratory of Navigation and Location-based Services, School of Electronic Information and Electrical Engineering, Shanghai Jiao Tong University, for their constructive suggestions in preparation of the paper. The authors would like to thank Dr. Miaomiao Wang from Western University for his guidance in the proof of Theorem 1. The first author would also like to thank Gongmin Yan from Northwestern Polytechnical University for providing the ring laser gyroscope based test data.

## Appendix A

The proof of **Theorem 1** is given as follows.

According to (51), $\mathbf{T}_1\mathbf{T}_2$ can be calculated as

$$\begin{aligned}\mathbf{T}_1\mathbf{T}_2 &= \begin{bmatrix} \mathbf{C}_{b,1}^i & \mathbf{v}_1^i & \mathbf{r}_1^i \\ 0_{1\times3} & 1 & 0 \\ 0_{1\times3} & 0 & 1 \end{bmatrix}\begin{bmatrix} \mathbf{C}_{b,2}^i & \mathbf{v}_2^i & \mathbf{r}_2^i \\ 0_{1\times3} & 1 & 0 \\ 0_{1\times3} & 0 & 1 \end{bmatrix} \\ &= \begin{bmatrix} \mathbf{C}_{b,1}^i\mathbf{C}_{b,2}^i & \mathbf{C}_{b,1}^i\mathbf{v}_2^i + \mathbf{v}_1^i & \mathbf{C}_{b,1}^i\mathbf{r}_2^i + \mathbf{r}_1^i \\ 0_{1\times3} & 1 & 0 \\ 0_{1\times3} & 0 & 1 \end{bmatrix}\end{aligned} \tag{A1}$$

Then $\Gamma_{\mathbf{u}}\left(\mathbf{T}_1\mathbf{T}_2\right)$ is given by

$$\begin{aligned}&\Gamma_{\mathbf{u}}\left(\mathbf{T}_1\mathbf{T}_2\right) \\ &= \begin{bmatrix} \mathbf{C}_{b,1}^i\mathbf{C}_{b,2}^i\left(\boldsymbol{\omega}_{ib}^b\times\right) & \mathbf{C}_{b,1}^i\mathbf{C}_{b,2}^i\mathbf{f}^b + \mathbf{C}_n^i\mathbf{g}^n & \mathbf{C}_{b,1}^i\mathbf{v}_2^i + \mathbf{v}_1^i \\ 0_{1\times3} & 0 & 0 \\ 0_{1\times3} & 0 & 0 \end{bmatrix}\end{aligned} \tag{A2}$$

$\Gamma_{\mathbf{u}}\left(\mathbf{T}_1\right)\mathbf{T}_2$ is given by

$$\begin{aligned}\Gamma_{\mathbf{u}}\left(\mathbf{T}_1\right)\mathbf{T}_2 &= \begin{bmatrix} \mathbf{C}_{b,1}^i\boldsymbol{\omega}_{ib}^b\times & \mathbf{C}_{b,1}^i\mathbf{f}^b + \mathbf{C}_n^i\mathbf{g}^n & \mathbf{v}_1^i \\ 0_{3\times1} & 0 & 0 \\ 0_{3\times1} & 0 & 0 \end{bmatrix}\begin{bmatrix} \mathbf{C}_{b,2}^i & \mathbf{v}_2^i & \mathbf{r}_2^i \\ 0_{3\times1} & 1 & 0 \\ 0_{3\times1} & 0 & 1 \end{bmatrix} \\ &= \begin{bmatrix} \mathbf{C}_{b,1}^i\left(\boldsymbol{\omega}_{ib}^b\times\right)\mathbf{C}_{b,2}^i & \begin{matrix}\mathbf{C}_{b,1}^i\left(\boldsymbol{\omega}_{ib}^b\times\right)\mathbf{v}_2^i \\ +\mathbf{C}_{b,1}^i\mathbf{f}^b + \mathbf{C}_n^i\mathbf{g}^n\end{matrix} & \mathbf{C}_{b,1}^i\left(\boldsymbol{\omega}_{ib}^b\times\right)\mathbf{r}_2^i + \mathbf{v}_1^i \\ 0_{3\times1} & 0 & 0 \\ 0_{3\times1} & 0 & 0 \end{bmatrix}\end{aligned} \tag{A3}$$

$\mathbf{T}_1\Gamma_{\mathbf{u}}\left(\mathbf{T}_2\right)$ is given by

$$\begin{aligned}\mathbf{T}_1\Gamma_{\mathbf{u}}\left(\mathbf{T}_2\right) &= \begin{bmatrix} \mathbf{C}_{b,1}^i & \mathbf{v}_1^i & \mathbf{r}_1^i \\ 0_{3\times1} & 1 & 0 \\ 0_{3\times1} & 0 & 1 \end{bmatrix}\begin{bmatrix} \mathbf{C}_{b,2}^i\boldsymbol{\omega}_{ib}^b\times & \mathbf{C}_{b,2}^i\mathbf{f}^b + \mathbf{C}_n^i\mathbf{g}^n & \mathbf{v}_2^i \\ 0_{3\times1} & 0 & 0 \\ 0_{3\times1} & 0 & 0 \end{bmatrix} \\ &= \begin{bmatrix} \mathbf{C}_{b,1}^i\mathbf{C}_{b,2}^i\left(\boldsymbol{\omega}_{ib}^b\times\right) & \mathbf{C}_{b,1}^i\left(\mathbf{C}_{b,2}^i\mathbf{f}^b + \mathbf{C}_n^i\mathbf{g}^n\right) & \mathbf{C}_{b,1}^i\mathbf{v}_2^i \\ 0_{3\times1} & 0 & 0 \\ 0_{3\times1} & 0 & 0 \end{bmatrix}\end{aligned} \tag{A4}$$

$\mathbf{T}_1\Gamma_{\mathbf{u}}\left(\mathbf{I}_{5\times5}\right)\mathbf{T}_2$ is given by

$$\begin{aligned}&\mathbf{T}_1\Gamma_{\mathbf{u}}\left(\mathbf{I}_{5\times5}\right)\mathbf{T}_2 = \\ &\begin{bmatrix} \mathbf{C}_{b,1}^i & \mathbf{v}_1^i & \mathbf{r}_1^i \\ 0_{1\times3} & 1 & 0 \\ 0_{1\times3} & 0 & 1 \end{bmatrix}\begin{bmatrix} \left(\boldsymbol{\omega}_{ib}^b\times\right) & \mathbf{f}^b + \mathbf{C}_n^i\mathbf{g}^n & 0_{3\times1} \\ 0_{1\times3} & 0 & 0 \\ 0_{1\times3} & 0 & 0 \end{bmatrix}\begin{bmatrix} \mathbf{C}_{b,2}^i & \mathbf{v}_2^i & \mathbf{r}_2^i \\ 0_{1\times3} & 1 & 0 \\ 0_{1\times3} & 0 & 1 \end{bmatrix} \\ &= \begin{bmatrix} \mathbf{C}_{b,1}^i\left(\boldsymbol{\omega}_{ib}^b\times\right)\mathbf{C}_{b,2}^i & \begin{matrix}\mathbf{C}_{b,1}^i\left(\boldsymbol{\omega}_{ib}^b\times\right)\mathbf{v}_2^i \\ +\mathbf{C}_{b,1}^i\left(\mathbf{f}^b + \mathbf{C}_n^i\mathbf{g}^n\right)\end{matrix} & \mathbf{C}_{b,1}^i\left(\boldsymbol{\omega}_{ib}^b\times\right)\mathbf{r}_2^i \\ 0_{1\times3} & 0 & 0 \\ 0_{1\times3} & 0 & 0 \end{bmatrix}\end{aligned} \tag{A5}$$

According to (A2)-(A5), it can be easily demonstrated that

$$\Gamma_{\mathbf{u}}\left(\mathbf{T}_1\mathbf{T}_2\right) = \Gamma_{\mathbf{u}}\left(\mathbf{T}_1\right)\mathbf{T}_2 + \mathbf{T}_1\Gamma_{\mathbf{u}}\left(\mathbf{T}_2\right) - \mathbf{T}_1\Gamma_{\mathbf{u}}\left(\mathbf{I}_{5\times5}\right)\mathbf{T}_2 \tag{A6}$$

10.1109/TAES.2020.3011998, 2020.